\documentclass[sigconf, manuscript, screen,nonacm]{acmart}
\AtBeginDocument{%
  \providecommand\BibTeX{{%
    \normalfont B\kern-0.5em{\scshape i\kern-0.25em b}\kern-0.8em\TeX}}}

\usepackage[linesnumbered,boxed]{algorithm2e}
\usepackage[caption = false]{subfig}
\usepackage{float}
\usepackage{amsmath}

\usepackage{amsfonts}
\usepackage{amssymb}
\begin{document}

\title{Deep metric learning improves lab of origin prediction of genetically engineered plasmids}

\author{Fernando H. F. Camargo}
\authornote{Both authors contributed equally to this research.}
\email{fernando.camargo.ai@gmail.com}
\author{Igor Muniz Soares}
\authornotemark[1]
\email{igor@amalgam.ai / igor.muniz.ims@gmail.com}
\affiliation{%
  \institution{Amalgam}
  \streetaddress{11 N Skokie Hwy, Ste 201}
  \city{Lake Bluff}
  \state{Illinois}
  \country{USA}
  \postcode{60044}
}

\author{Adriano M. Marques}
\email{adriano@amalgam.ai}
\affiliation{%
  \institution{Amalgam}
  \streetaddress{11 N Skokie Hwy, Ste 201}
  \city{Lake Bluff}
  \state{Illinois}
  \country{USA}
  \postcode{60044}
}

\author{Oliver M. Crook}
\email{oliver.crook@stats.ox.ac.uk}
\affiliation{%
  \institution{Oxford Protein Informatics Group}
  \streetaddress{University of Oxford}
  \city{Oxford}
  \state{}
  \country{UK}
  \postcode{}
}

\renewcommand{\shortauthors}{Camargo and Soares, et al.}

\begin{abstract}
Genome engineering is undergoing unprecedented
development and is now becoming widely available. To ensure responsible biotechnology innovation and to reduce misuse of engineered DNA sequences, it is vital to develop tools to identify the lab-of-origin of engineered plasmids. Genetic engineering attribution (GEA), the ability to make sequence-lab associations, would support forensic experts in this process. Here, we propose a method, based on metric learning, that ranks the most likely labs-of-origin whilst simultaneously generating embeddings for plasmid sequences and labs. These embeddings can be used to perform various downstream tasks, such as clustering DNA sequences and labs, as well as using them as features in machine learning models. Our approach employs a circular shift augmentation approach and is able to correctly rank the lab-of-origin $90\%$ of the time within its top 10 predictions - outperforming all current state-of-the-art approaches. We also demonstrate that we can perform few-shot-learning and obtain $76\%$ top-10 accuracy using only $10\%$ of the sequences. This means, we outperform the previous CNN approach using only one-tenth of the data. We also demonstrate that we are able to extract key signatures in plasmid sequences for particular labs, allowing for an interpretable examination of the model's outputs.

\end{abstract}




\maketitle

\section{Introduction}\label{sec:intro}
Genetic engineering and synthetic biology are fast growing areas of biotechnology. We are now able to transform organisms in highly efficient and sophisticated manners. As this biotechnology becomes more widespread, it is vital that we may attribute genetically engineered organisms to their makers or lab-of-origin. This will prevent plagiarism, encourage responsible development, allow designers to gain due credit and as an approach to holding genetic engineers/designers accountable for their work. Tools for attributing this biotechnology to their owners, often referred to \textit{genetic engineering attribution} (GEA), have only recently become sufficiently well-performing \citep{Alley2020.08.22.262576, Nielsen2018, Wang2021}. When making design choices for nucleic-acid sequences, an engineer will impart a design signature which could be detectable by GEA methods. Powerful methods could, in principle, identify the true designer of a biological sequence and hence be excellent tools for accountability.   

Several approaches to GEA have now been proposed based on predicting the lab-of-origin of plasmid sequences from the Addgene \cite{Addgene} data repository. The performance of these approaches has quickly improved, rising from $70\%$ top-10 accuracy \citep{Nielsen2018} to $85\%$ top-10 accuracy \citep{Alley2020, Wang2021} in recent years. \citet{Nielsen2018} used Convolutional Neural Networks (CNNs), \citet{Alley2020} used Recurrent Neural Networks (RNNs), and \citet{Wang2021} used a pan-genome approach, suggesting a variety of methods could perform well on this task. However, further improvements are still possible, these approach make other downstream tasks challenging and require many training instances to perform well. Here, we present an approach that couples CNNs and metric learning \cite{kulis2013} to extract embeddings from DNA sequences, whilst simultaneously learning an embedding of known labs. We also employ circular shift augmentation rather than the typically used reverse-complement augmentation. Together, our method improves over the state-of-the-art by 5 percentage points and allows clustering of sequence and labs for other downstream tasks. Our approach allows us to perform one-shot-learning \cite{one-shot} and so we can predict lab associations with only 1 training instance. Furthermore, using integrated-gradients we can extract design signatures from our model, allowing us to interpret the model outputs.

\begin{figure*}[htp]
  \centering
\includegraphics[width=\linewidth]{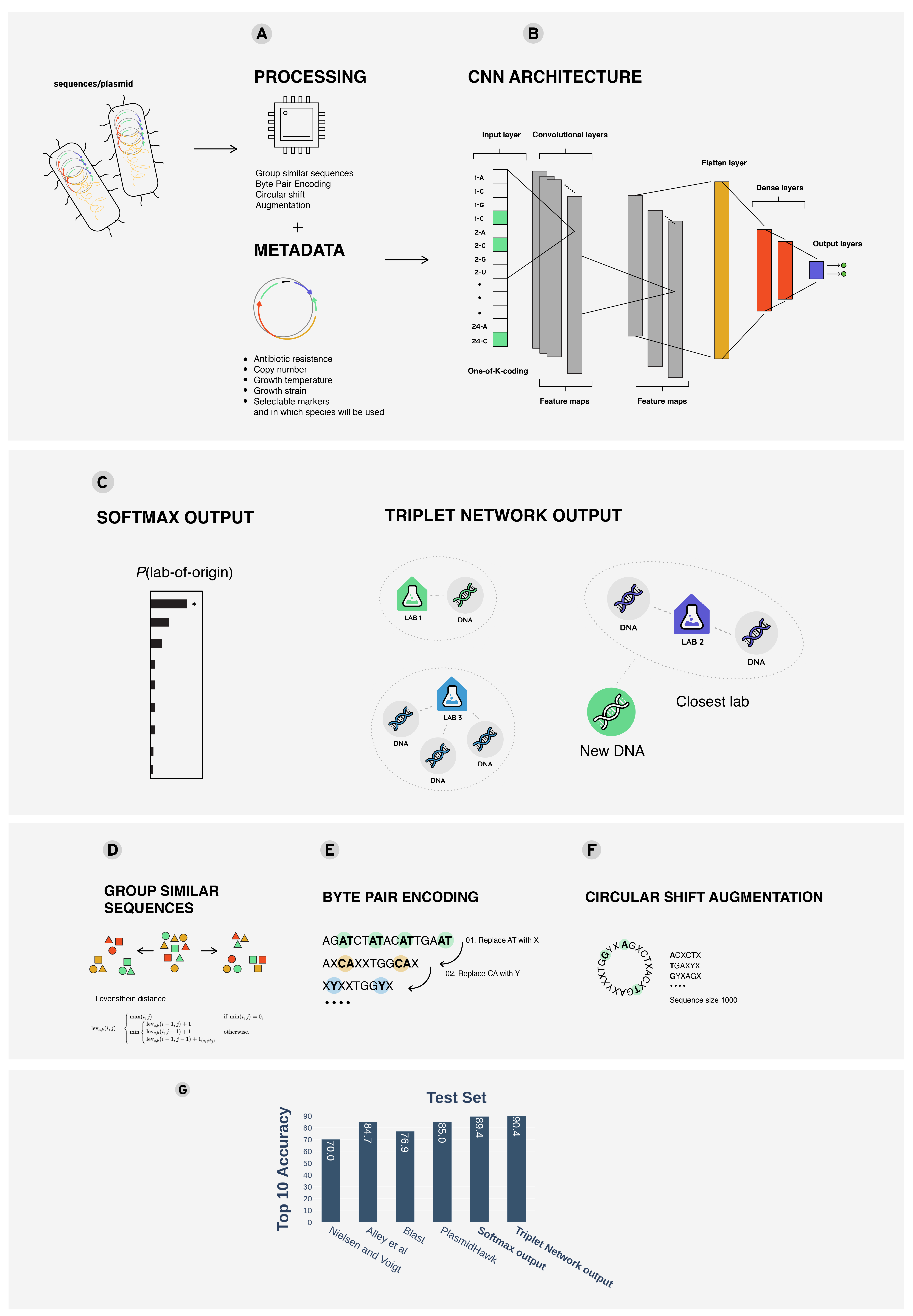}
  \caption{Applying Convolutional Neural Networks with Metric Learning on plasmids leads to a new state-of-the-art.}
  \label{fig:arch}
\end{figure*}

\begin{figure}[htp]
  {\textbf{a} Genetic sequences and phenotype information are input into our model. All sequences are processed in order to better demonstrate the characteristics of each plasmid and improve the model's ability to identify patterns.  \textbf{b} We use convolutional neural networks as the base model for our two approaches. In this model, convolutional operations extract information present in the sequence based on a fixed kernel size. Our method uses convolutional structures with different kernel sizes in parallel, simulating the observation of sequences by pieces of different sizes. The information extracted by each structure is aggregated and added to the phenotype information and then assigned to one of the laboratories. \textbf{c} The difference between the two approaches is the training and output of the model. The standard approach treats the genetic attribution problem as a classification problem where a softmax layer is applied to determine the probability that the sequence belongs to each of the laboratories seen by the model in the training phase. On the other hand, a metric learning approach (triplet network) determines how far the features representation of a new sequence is from the sequences cluster of a laboratory in the base. Smaller distances indicate greater similarities between the features of a sequence and a lab-of-origin. \textbf{d} Before training, all sequences from a given laboratory are grouped according to their Levenshtein distances. We do not use sequences from the same group in training and validation at the same time, ensuring that sequences that are too close do not cause leakage in training and overfit the model. \textbf{e} DNA sequences are compressed by Byte Pair Encoding (BPE) algorithm \cite{gage1994}. It works by looking for common patterns in the sequence and unifying them into tokens, increasing the vocabulary while reducing the sequences' size. \textbf{f} Since plasmids are circular sequences,  we randomly shift the starting point of the sequence, increasing our number of training data. This method is performed "online" during sequence loading and preparing for network entry. After all these processing, we limited the sequence size to 1000 characters to optimize the training time and convergence capability of the algorithm. \textbf{g} Top-10 prediction accuracy on the test set. We compared our both approaches to \citeauthor{Nielsen2018} , \citeauthor{Alley2020}, BLAST baseline \cite{Alley2020} and PlasmidHawk \cite{Wang2021}.}
\end{figure}

We evaluate our approach on a dataset from the Addgene repository containing 81,834 DNA sequences along with minimal phenotypic information: antibiotic resistance, copy number, growth temperature, growth strain, selectable markers and in which species (see methods). The plasmids are designed by 3,751 different labs and grouped into 1,313 categories, along with a single additional category to represent "unknown engineered" (see methods). We evaluate the solutions using accuracy and top-10 accuracy metric. Top-10 accuracy means that the model needs to rank the correct lab-of-origin within the ten most likely labs. Ranking often requires a slightly different approach to classification, and so we developed a Metric Learning approach \cite{kulis2013} (more specifically, Triplet Networks \cite{hoffer2018deep}). We begin the manuscript by outlining our method, demonstrating that it improves over the state-of-the-art. We then demonstrate that the embedding approach allows us to perform other tasks of interest and finally demonstrate how we can perform one/few-shot-learning \cite{one-shot, NIPS2004_ef1e491a, LiFeiFei2006, Wang2020} with our approach.

\section{Results}


\subsection{Metric learning model and model evaluation}

Our proposed method uses (deep) metric learning \cite{kulis2013}, which is where one learns a distance function between objects. Here, this can be thought of learning a similarity between plasmid sequences and labs. The result is an embedding where distances between sequences have their similarity preserved. We use deep learning, in particular a CNN-based approach, to extract embeddings of DNA sequences, while learning the embeddings of the known labs. To demonstrate that using deep metric learning indeed provides an advantage, we also developed a regular classifier with a similar architecture to compare with our deep metric learning approach. 

Each model is composed of a CNN with multiple kernels of differing sizes \cite{kim2014}. The CNN is used to extract features from the sequence, which are then concatenated with the phenotypic metadata of that plasmid sequence. The key difference between our two proposed models can be understood by considering the final layers. The classification model's final layer has a softmax activation function, resulting in each lab being given a probability vector associating it with each lab. Instead, our metric learning approach passes these features through a dense layer that generates our sequence embedding. In parallel, we have an embedding layer that learns the lab embeddings. The principle advantage of our metric learning approach is that once it has been trained it can extract embeddings of any DNA sequence, allowing us to group or cluster new sequences to existing labs based on similar characteristics.

Our classification model is trained using regular supervised learning. However, our metric learning approach is trained differently. Specifically, our metric learning method employs the idea of Triplet Networks \citep{hoffer2018deep}. Here, we create a triplet (anchors, positives, negatives) as part of the model training process. Our model is anchored around the DNA sequences and hence they are the anchors in this approach. The positive object in this scenario is the true lab-of-origin, whilst the negative object is some other lab. To be concrete suppose $s_1$ is a plasmid sequence made by the Church lab, then a possible triplet would be $(s_1, \text{Church lab}, \text{Voigt lab})$. Hence, the goal of our approach is to generate embeddings in which the DNA sequences are near their labs-of-origin and far from the sequences of other labs. Using the Addgene dataset (see methods), we trained each approach to perform DNA sequence prediction to one of $1,313$ laboratories or "unknown engineered" (see methods). The dataset has a total of $81,834$ DNA sequences, where $18,817$ of these sequences were separated for testing. During training, we split the data into $85\%$ for algorithm training and $15\%$ for validation. Our complete method can be summarised in Figure \ref{fig:arch}.

\begin{figure*}[t]
  \centering
\includegraphics[width=\linewidth]{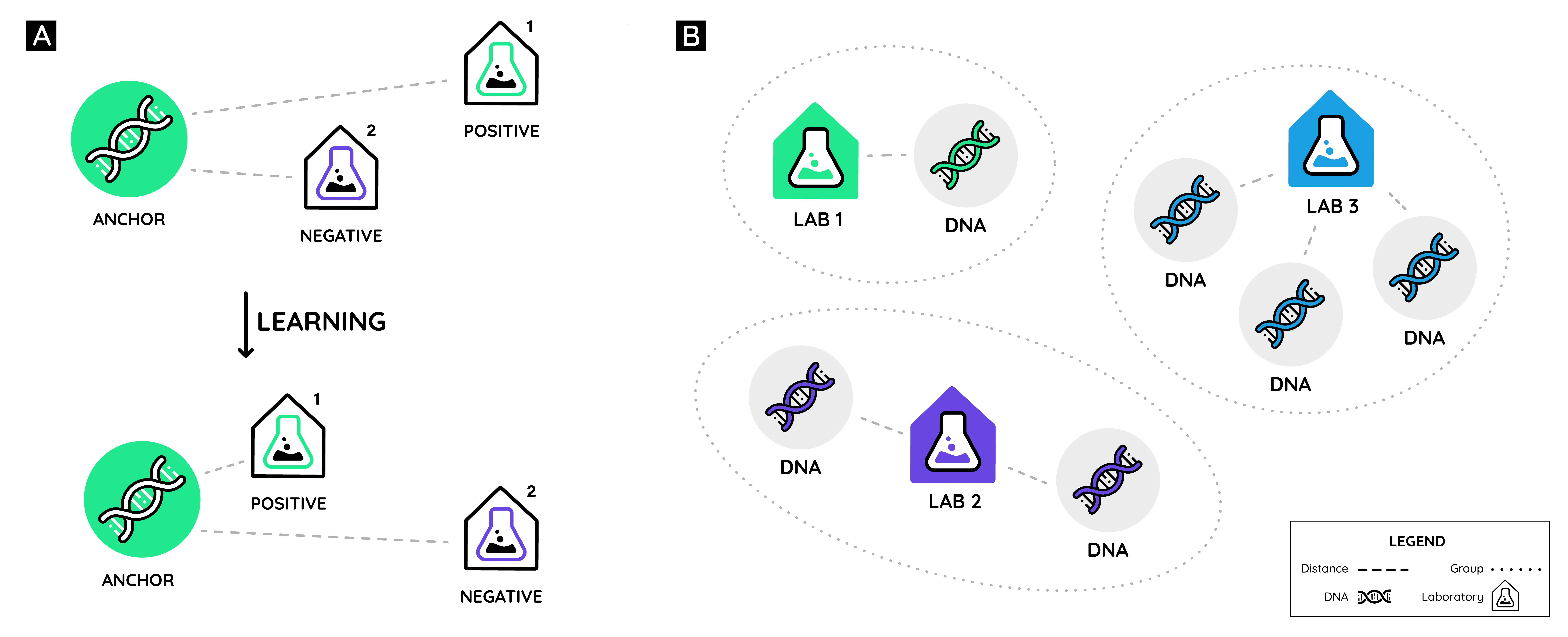}
  \caption{The triplet is composed of an anchor (DNA), a positive (the lab-of-origin), and a negative example (another lab). \textbf{a} In the beginning, the anchor might be closer to the negative than it is to the positive. During training, we pull the anchor and positive towards each other while pushing the negative away. \textbf{b} In the end, labs and their DNA sequences will be nearer to each other, forming groups. We can also expect both labs and DNA sequences to be closer to other similar ones.}
  \label{fig:triplet}
\end{figure*}



\subsection{Metric learning model improves predictions over state-of-the-art}

\citet{Nielsen2018} developed a deep learning model by applying convolutional neural networks. The network was trained on the Addgene plasmid dataset and independently verified in \citep{Alley2020}. In brief, their approach can be summarised as follows. First, DNA sequences were one-hot encoded, which were then used as input to the network composed of one convolutional layer of 128 filters, max-pooling operation, and two dense layers. Whilst, they showed it is possible to use machine learning for this task, this seminal approach was hopeful and obtained an accuracy of only $48\%$ and a top-10 accuracy of $70\%$ in predicting the origin lab. 

More recently, \citet{Alley2020} proposed deteRNNt, a recurrent neural network-based model. The main insight of this approach was to treat the DNA sequence as a text problem, using techniques from the natural language processing field to extract features from the sequence. They tokenized the sequence using Byte Pair Encoding \cite{gage1994}, generating larger tokens and decreasing the size of the sequence. These tokens then served as input to a word embedding layer \cite{wordembedding} followed by recurrent neural networks \cite{rnn1} \cite{rnn2}. The authors showed that their approach achieves $84.7\%$ top-10 accuracy. 

We also compared with BLAST \cite{Altschul1990}, with our test set of 18,817 samples. Despite being a relatively simple tool that employs no modern machine learning and simply finds similar local regions between sequences, it was able to predict source labs by achieving $76.9\%$ top-10 accuracy in our tests: outperforming the approach of \citet{Nielsen2018}

PlasmidHawk \cite{Wang2021}, a recently launched tool, uses Plaster \citep{Wang::2019}, a state-of-the-art pan-genome algorithm to construct a synthetic plasmid resulting in a set of sequence fragments. It then aligns the original plasmid to the synthetic one and makes comparisons to match fragments with the plasmid. This method has so far outperformed other machine learning-based methods by obtaining $85\%$ of top-10 accuracy, whilst employing no machine learning.

Here, we find that our metric learning model alongside our training methodology improves the current state-of-the-art for attributing the lab-of-origin of an engineered DNA sequence, achieving $90.39\%$ top-10 accuracy. For the classic approach of a classification model that predicts the input sequence's probability from any of the possible labs seen during training, our methodology also surpasses all previous methods reaching $89.36\%$ top-10 accuracy. These methods represent a $4-5$ absolute percentage improvement in performance over the current state-of-the-art. Whilst our metric learning approach improves over a simple softmax-based method using similar CNN architecture by $1$ percentage point. 


\subsection{Using triplet networks to embed the DNA sequences}

We train Triplet Networks to learn embeddings (vector representations with preserved distance) for both labs and DNA sequences. These vectors live in the same vector space, which allows us to compare them in a variety of ways. For example, we can compare: the distance between two labs; between a lab and a sequence; between a sequence and another sequence. This allows us to perform tasks other than ranking the possible origin lab of a given DNA sequence.

\begin{figure*}[h]
  \centering
  \subfloat[Distortion Score Elbow for KMeans Clustering]{\includegraphics[width=0.5\linewidth]{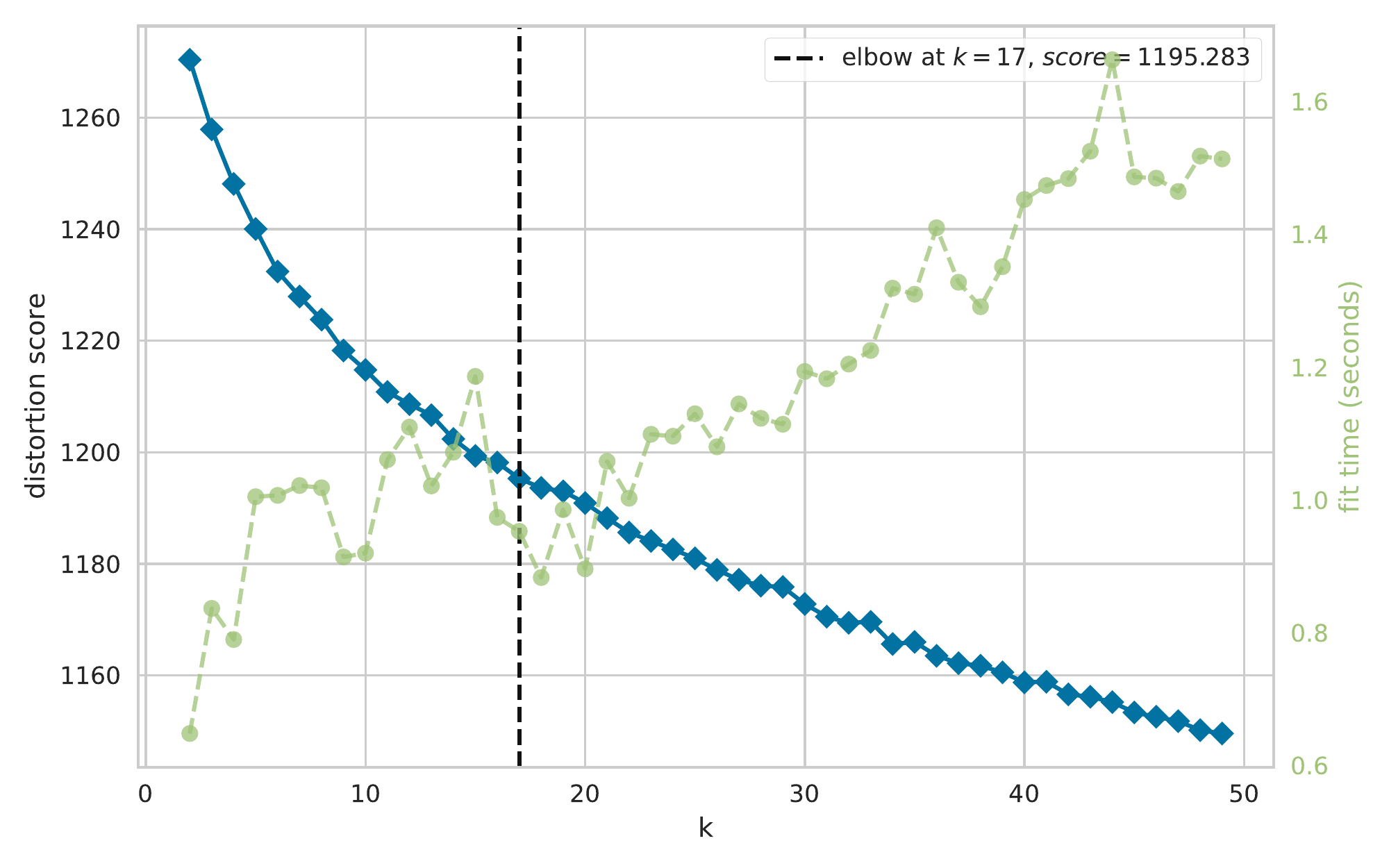}}
  \subfloat[Hierarchical Clustering Dendrogram (4 levels)]{\includegraphics[width=0.5\linewidth]{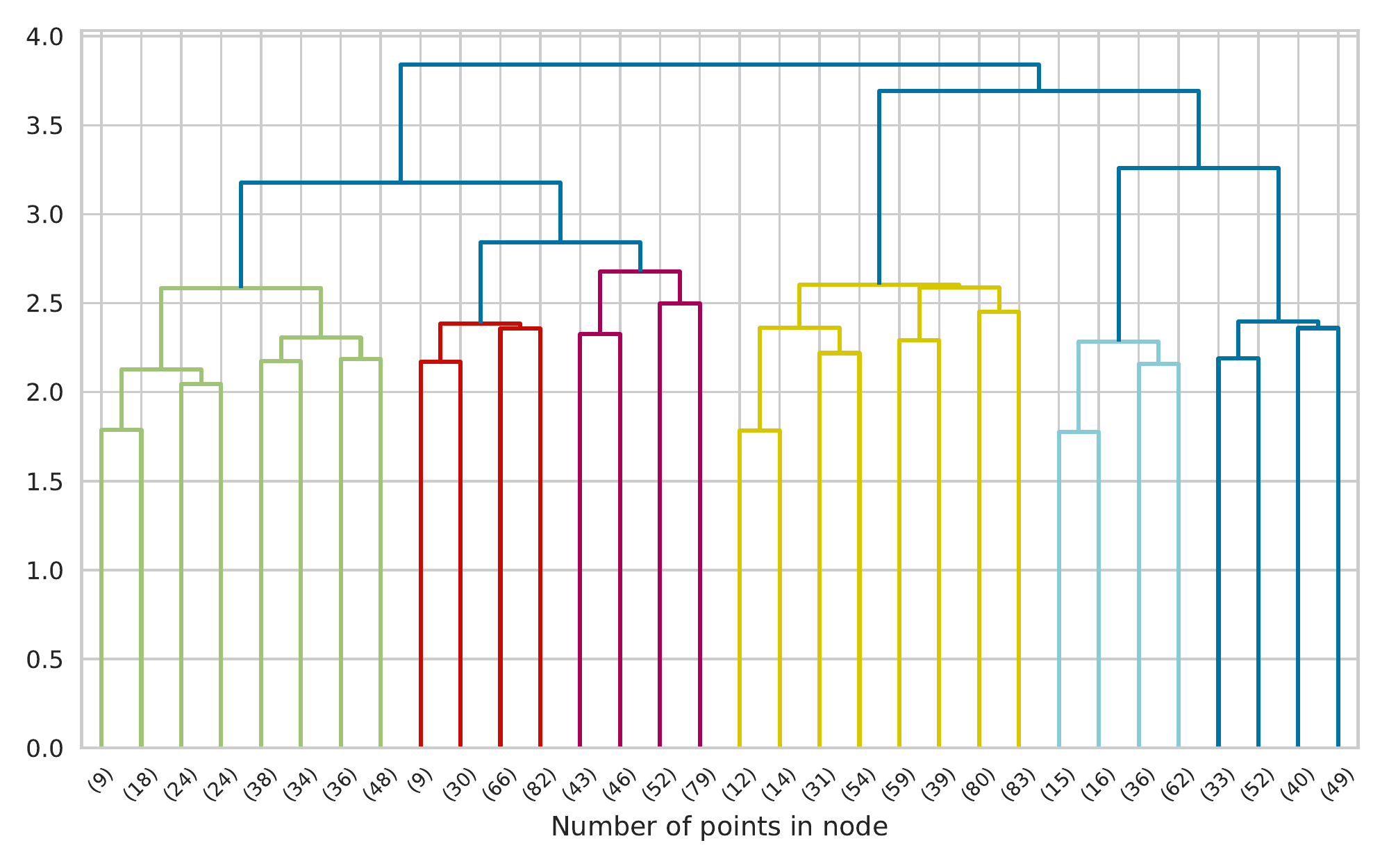}}\\
  \subfloat[Number of labs per cluster (17 clusters)]{\includegraphics[width=0.5\linewidth]{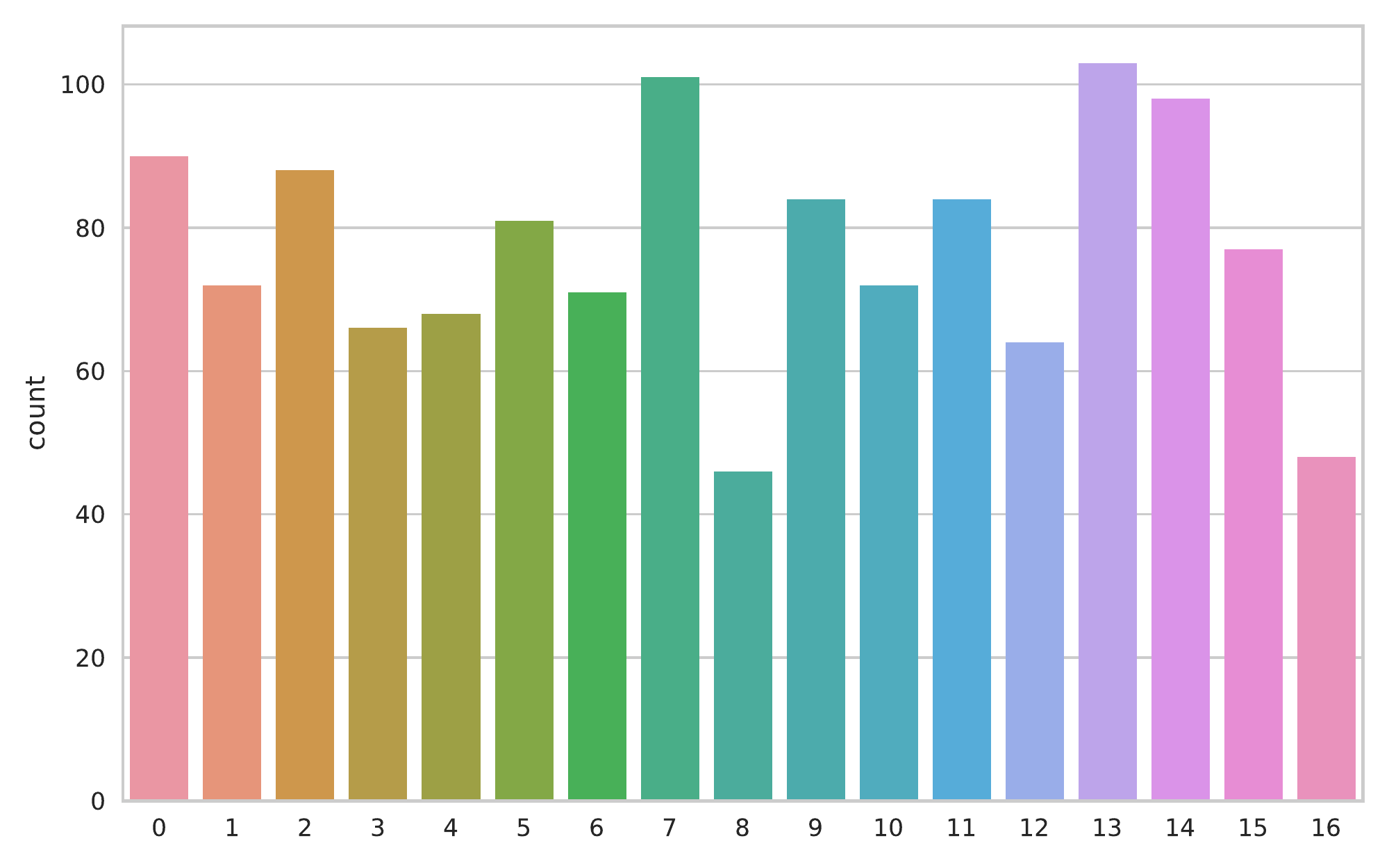}}
  \subfloat[Labs by cluster (17 clusters)
  ]{\includegraphics[width=0.5\linewidth]{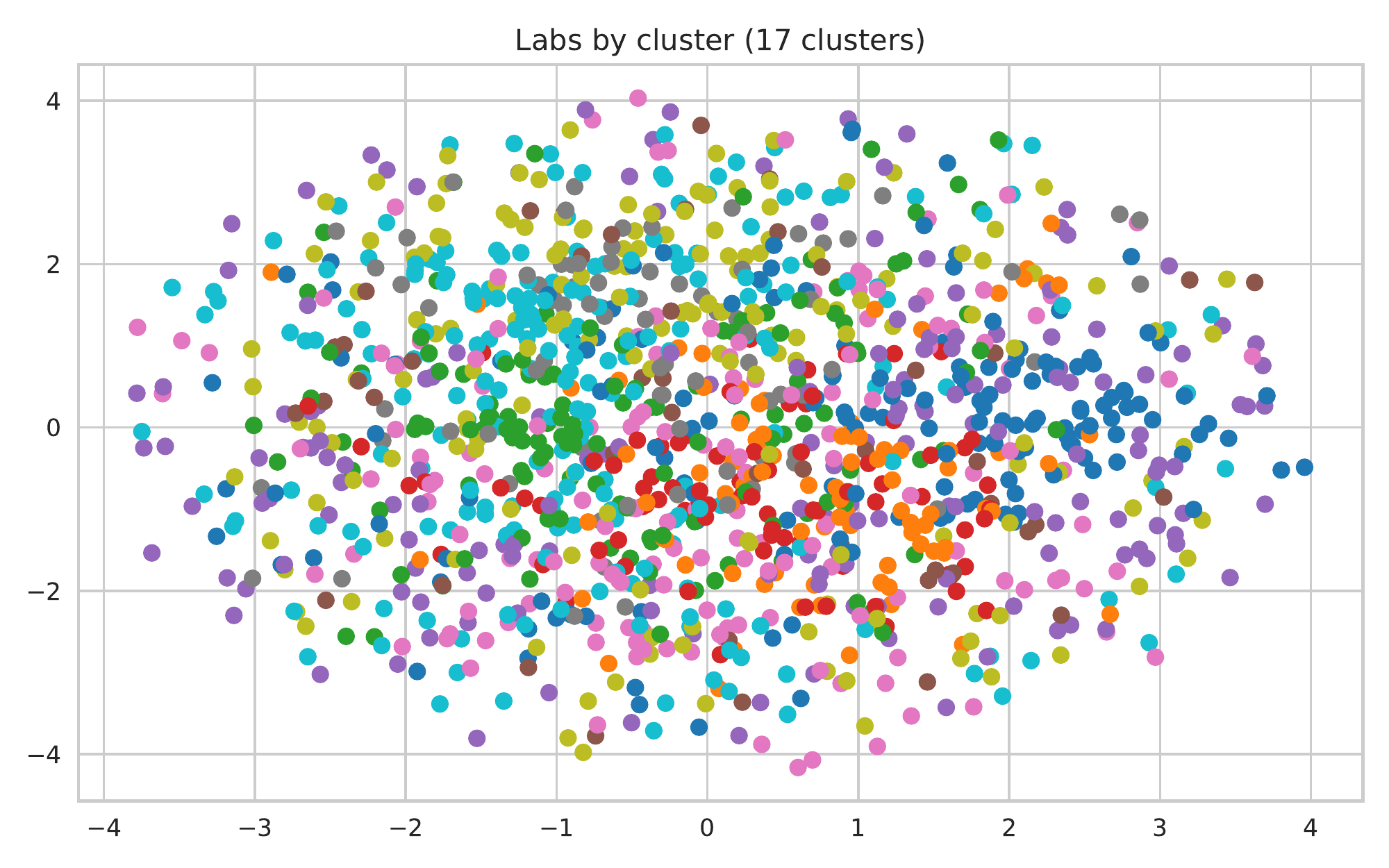}}
  \caption{\textbf{a} Shows the application of the Elbow Method, a commonly used method to find the best number of clusters. As we can see, we could group the labs with various numbers of clusters, but 17 seems to be the optimal number (with a slight margin). \textbf{b} Displays a Hierarchical Clustering Dendrogram, making it possible to see the different points to split clusters further. We can notice that some labs can be easily differentiated, while some are very similar according to euclidian distance. \textbf{c} Using the optimal number of 17 clusters, we have the number of labs per cluster. In general, the clusters are very similar in size, but some of them have a significant difference from the average. \textbf{d} Shows the labs in a 2D space (after compressing information using t-SNE \cite{hinton2002}). The colors represent their clusters.}
  \label{fig:lab_clustering}
\end{figure*}

Clustering is one common application that can provide insights \cite{clustering, clustering2, clustering3}. Many labs will share information about design techniques, will have been mentored or trained in another lab or will be directly collaborating. However, those relationships and similarities are not always known to us. Even though these similarities are not directly apparent through embeddings themselves, we can also examine which DNA sequences of a particular lab are more similar to those of other labs. Figure \ref{fig:lab_clustering} showcases how labs can be clustered with our model and Figure \ref{fig:tsne} shows labs and their designed DNA sequences. We observe heterogeneous sizes of clusters with some labs generating only similar DNA sequences (evidenced by compact homogenous clusters), whilst some are highly dispersed. 

\subsection{Few shot learning}
Machine learning algorithms typically require large quantities of training data, and in many applications, this makes it challenging when new classes are added. Sparse training data for these classes can result in poor quality predictions. In the case of GEA, some new labs may only have a single or few appropriate training instances. Training machine learning algorithms to perform well in this task is known as Few-Shot Learning (FSL) \cite{Wang2020, LiFeiFei2006, NIPS2004_ef1e491a}. Based on knowledge already acquired by a model trained in a similar task, the few-shot learning method can generalize to a new task using simply a few samples. Our proposed method can straightforwardly be adapted to the FSL situation. Since embeddings are the representation of features, it is possible to use them as the input for other machine learning algorithms or to calculate approximations between the embedding of a new sequence and all previously observed sequences. Thus, in a scenario of new laboratories with few sequences, we can store these few samples and perform the few-shot learning process by calculating the distance between an unknown sequence and the embeddings of these laboratories. It is sufficient that only one of the stored samples has characteristics similar to the unknown sequence for prediction to be made. 

\begin{figure*}[h]
  \centering
  \subfloat[Few-Shot Learning Top-10 Accuracy\label{fig:few_shot_accuracies}]{\includegraphics[width=0.5\textwidth]{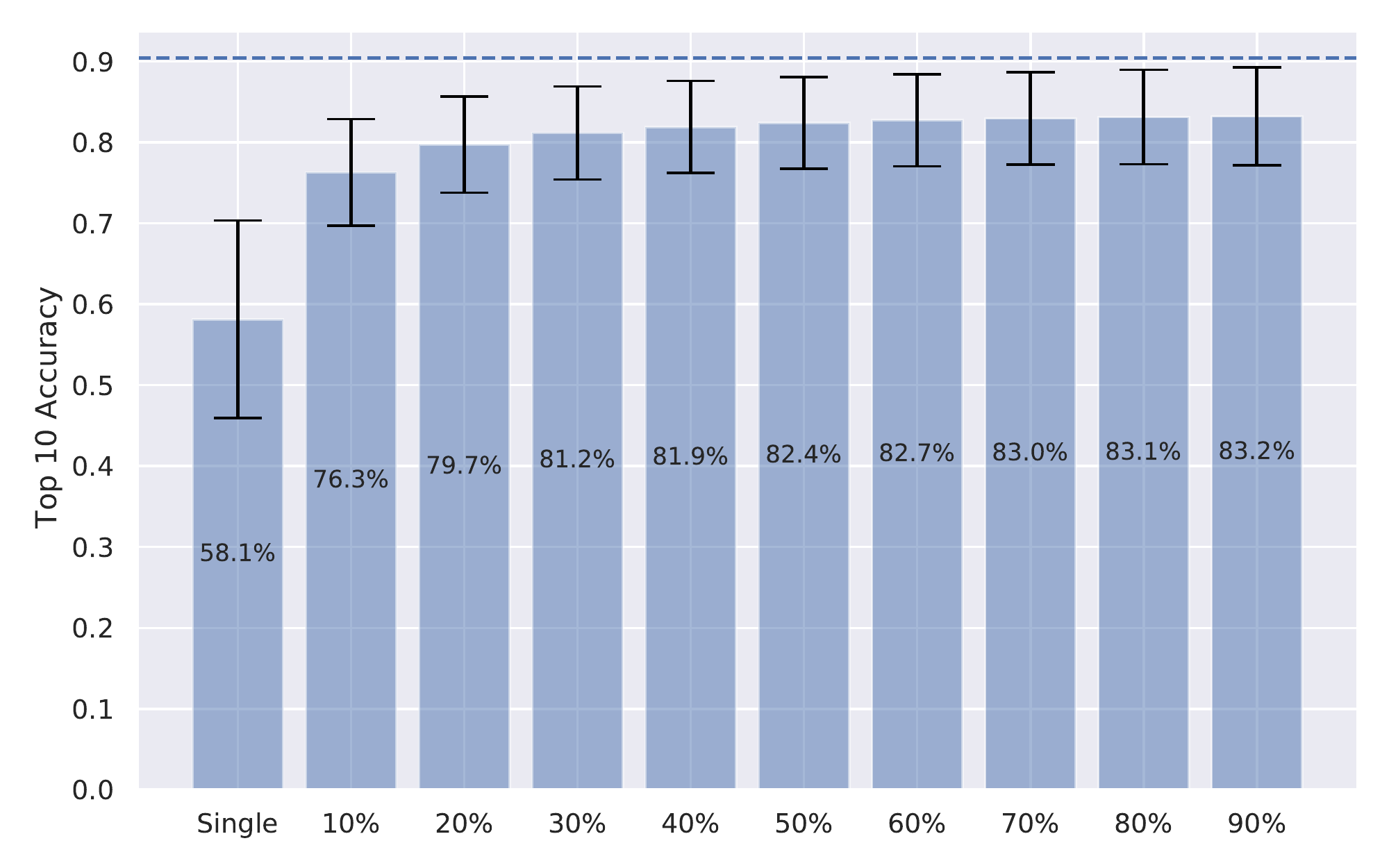}}
  \subfloat[Ranked position of correct lab using One-Shot Learning\label{fig:one_shot_positions}]{\includegraphics[width=0.5\textwidth]{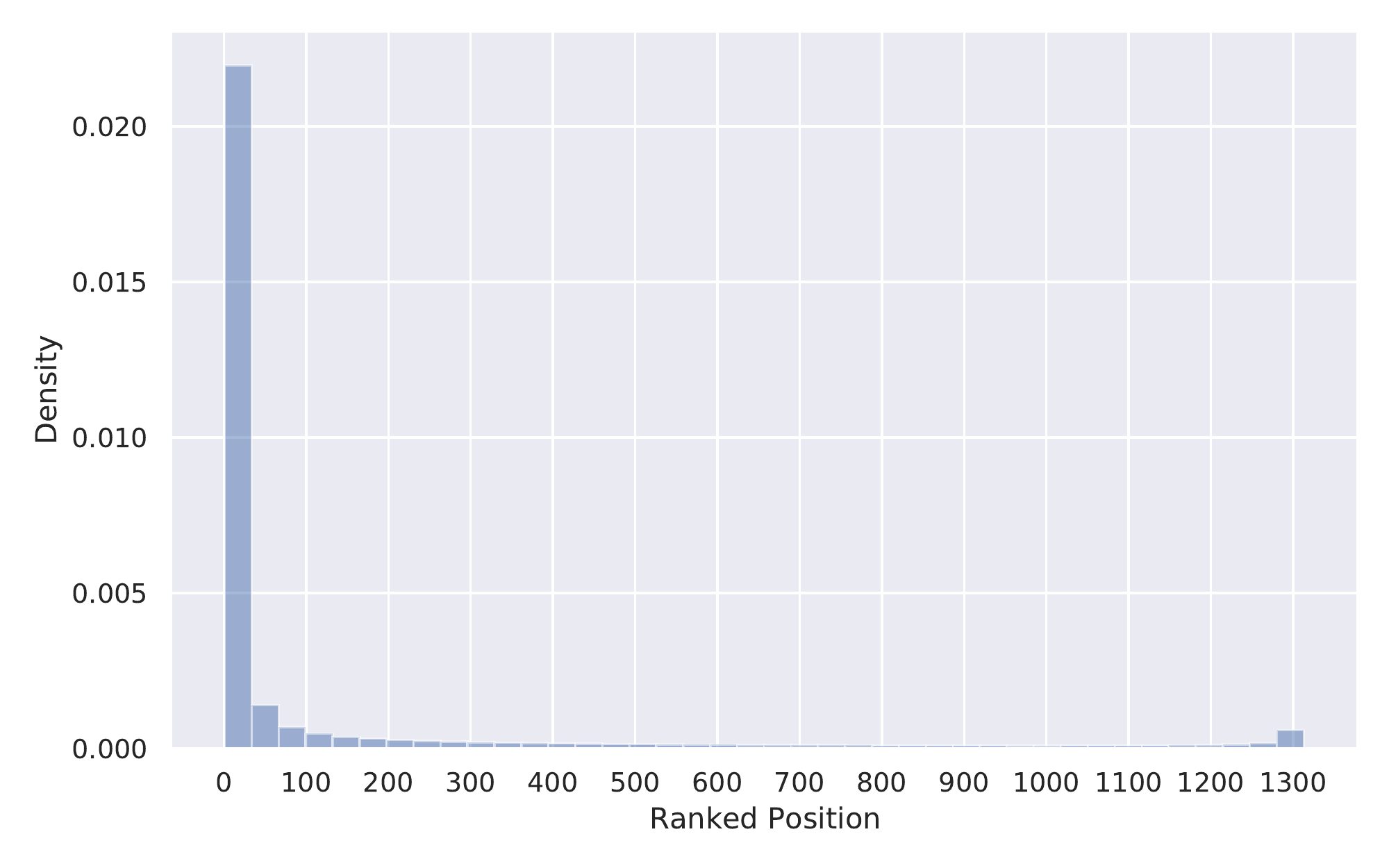}}
  \caption{\textbf{a} Each bar shows the mean and standard deviation of Top 10 Accuracy across all the experiments. It is worth noting that all the labs of the dataset had the chance to be left out of the training more than once with those experiments. The first bar refers to the extreme case where we pick a single plasmid to represent the lab. The other bars refer to picking a percentage of the plasmids to represent the lab and using the rest to evaluate. It is worth noting that the number of plasmids per lab varies a lot. There are labs in which 10\% will be one or two, while others that it will be hundreds of examples. For reference, the mean and standard deviation of each percentage are as follow: 10\% (5 +- 18), 20\% (9 +- 35), 30\% (13 +- 53), 40\% (17 +- 71), 50\% (22 +- 88), 60\% (26 +- 106), 70\% (31 +- 124), 80\% (35 +- 142), 90\% (39 +- 159). For last, the dashed line refers to our Top 10 Accuracy when retraining the model. \textbf{b} The histogram presents the ranked position of the lab-of-origin when using a single plasmid to represent it. As we can see, in most cases (79\%) a single plasmid is enough to rank it at least in the top 100 (given the labs in the dataset). It is worth noting that the median (50\% of the cases) ranked position was fifth. The variance can be high because there chosen plasmid may not be representative of the lab}
\end{figure*}

To test the ability of our approach to perform few-shot learning, we undertook the following experiment. We trained our method 100 times, each time we removed all the plasmids from 50 different labs in the training set. For each lab that was left out, we picked a random sample of plasmids to generate the embeddings that represent that lab. All remaining plasmids are used to evaluate our model. In the extreme case (also known as One-Shot Learning), we used a single plasmid to represent each lab and test with all the others. Figure \ref{fig:few_shot_accuracies} shows the metrics' mean and standard deviation of our model's capability with different sample sizes used to represent each lab. As we can see, the larger the sample we have, the greater the top-10 accuracy. However, there are diminishing returns as we increase the sample size. It shows that, in general, only a few representative examples are needed for high accuracy predictions. We see that our approach obtains better top-10 accuracy than the previously published CNN approach of \citet{Nielsen2018} whilst only using $10\%$ of the training data. Figure \ref{fig:one_shot_positions} and Table \ref{tab:one_shot_positions} show the rank of the lab-of-origin when taking a single plasmid to represent each lab. We observe that if we want to be sure the true lab is within our selected sample with probability $0.9$ then we need to include around $685$ labs. In other words, we can rule out around $50\%$ of the labs as the origin with confidence of $90\%$ when there was only a single plasmid available for that lab. Furthermore, it demonstrates that if an analyst can pick a single representative plasmid of a lab, using domain knowledge, then our approach has a good chance to attribute an unknown plasmid to that lab without any need to retrain the model.

\begin{table}[htp]
\centering
\begin{tabular}{ll}
\hline
Quantile & Position         \\ \hline
50\%     & 7 (top 0.4\%)    \\
60\%     & 17 (top 1.2\%)   \\
70\%     & 37 (top 2.7\%)   \\
80\%     & 180 (top 13.6\%) \\
90\%     & 685 (top 52\%)  
\end{tabular}
\caption{Given all the experiments, in which we pick a single plasmid to represent each lab, this Table shows the ranked position of the lab for each percentage of the cases. For example, 50\% of the time, the lab-of-origin was ranked 7th or lower. It is worth noting that the single plasmid for each lab was picked randomly and we do the same experiment multiple times. If an analyst could select a plasmid that they believe to be representative of the lab, we could expect an even better performance.}
\label{tab:one_shot_positions}
\end{table}

\subsection{Model interpretability and robustness}

Interpreting deep learning models gives us valuable information, such as understanding how the model works and the relative importance of features within the data. It can also reveal why some approaches work better than others, and this can be used to further improve the model. However, interpretation techniques for deep learning are still naive and is an area of active research \cite{Chakraborty2017InterpretabilityOD, doshivelez2017rigorous}.

In this work, we focus on understanding the differences between a triplet network and a conventional classification model, how robust our model is when performing point mutation, and most importantly, which tokens (parts of the sequence) are critical for identifying a lab.

\begin{figure}[h]
  \centering
  \subfloat[2D visualization tSNE Triplet Network]{\includegraphics[width=0.5\textwidth]{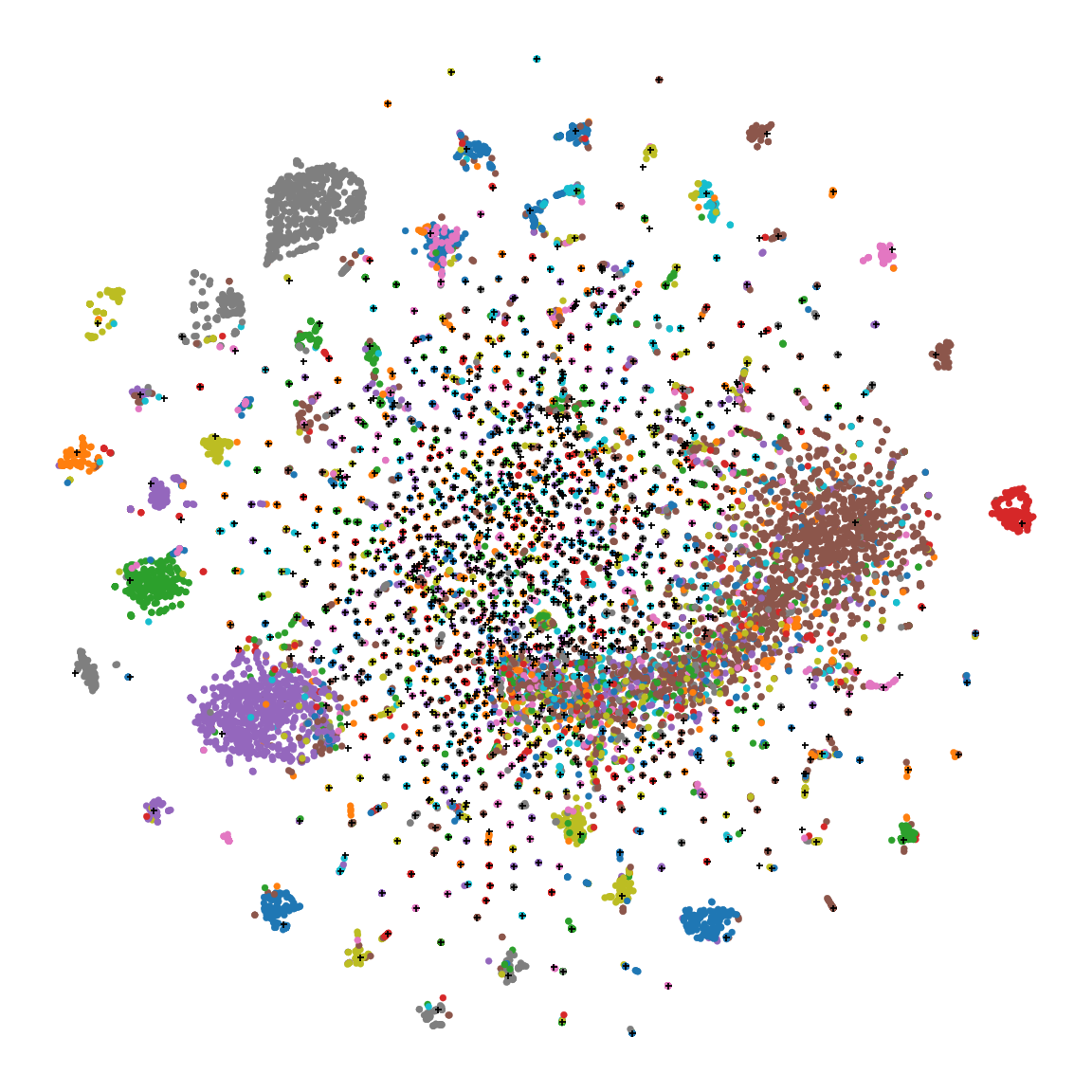}}
   \subfloat[2D visualization tSNE Softmax Classification]{\includegraphics[width=0.5\textwidth]{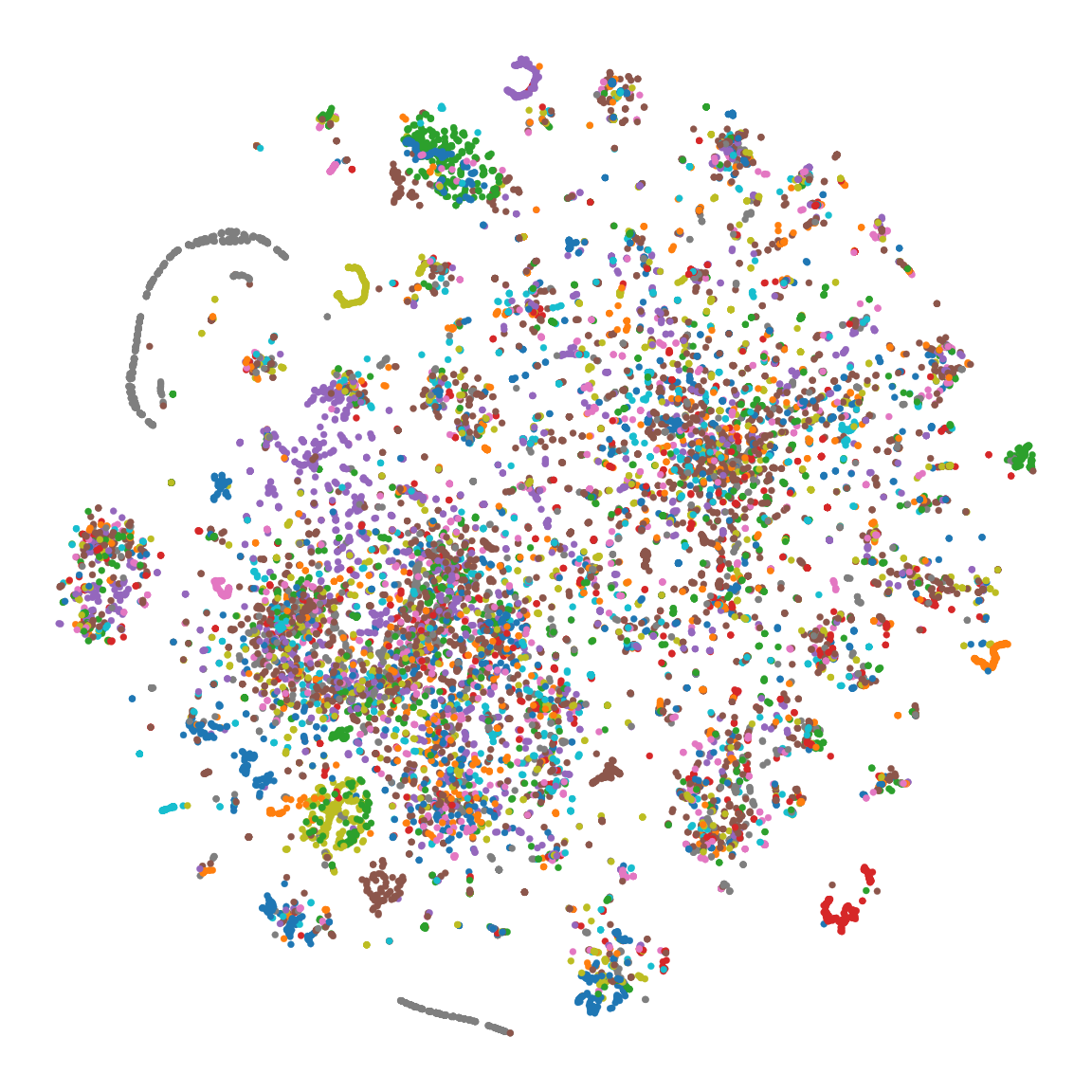}}\\
    \subfloat[Effect of point mutations on the Triplet model\label{fig:effect-mutations}]{\includegraphics[width=0.8\textwidth]{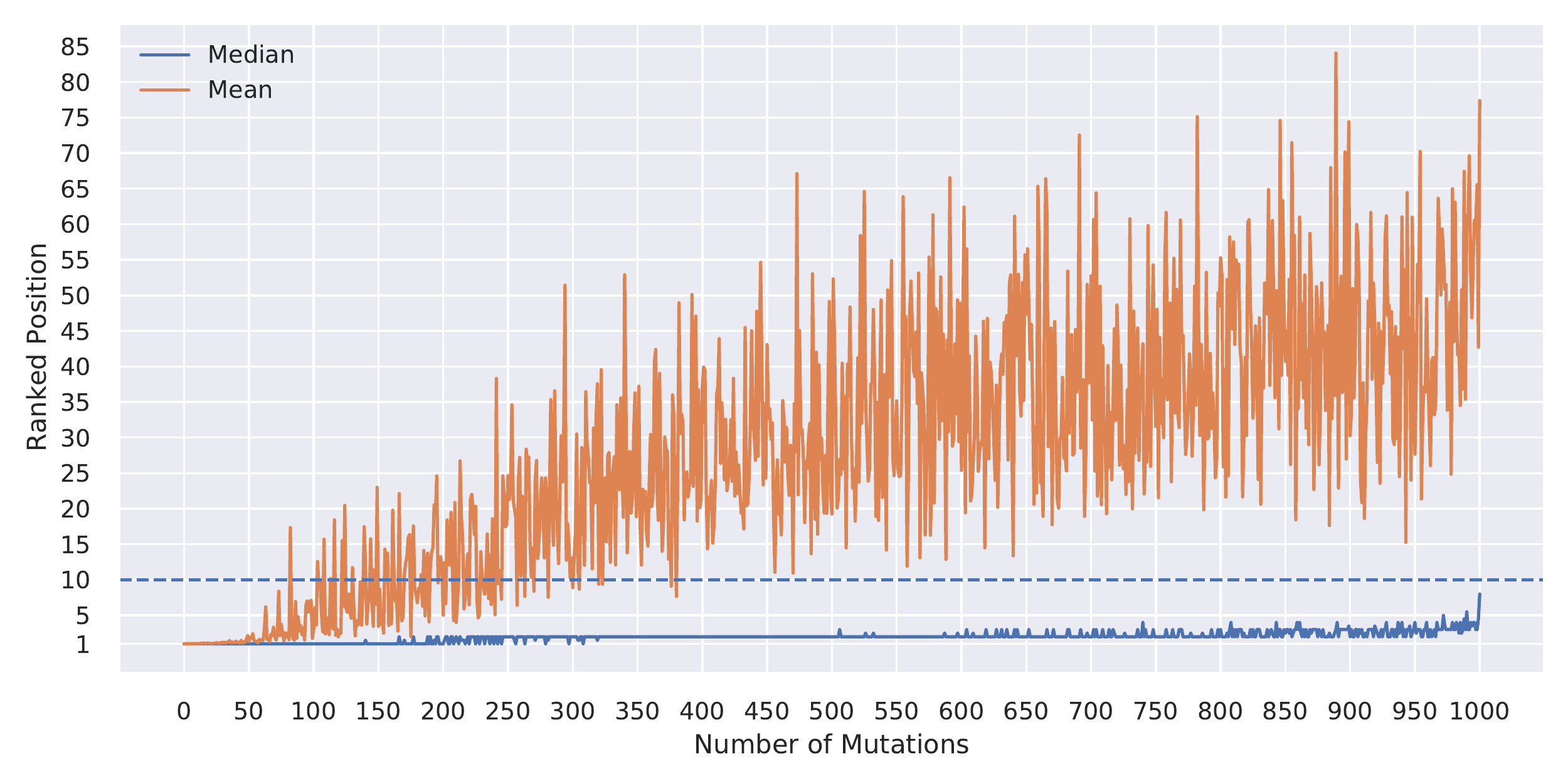}}
  \caption{\textbf{a} Each circle represents a DNA sequence, with its color highlighting its lab-of-origin. Each plus sign represents a lab. We project all of them from 200D to 2D using t-SNE for presentation purposes. We can see that the DNA sequences group together very well with their lab-of-origin. Some labs display very similar DNA sequences, while others are a bit dispersed. It showcases the differences between large and small labs. \textbf{b} tSNE visualization of the 3072D last hidden layer of the softmax model. Although we can see clusters in this feature map, they are not as well defined as in the triplet model. \textbf{c} Effect of point mutations on the Trilet model. The mean and meadian of 100 runs of the position of the correct lab in the model prediction ranking is shown, with the number of mutations ranging from 1 to 1000.}
  \label{fig:tsne}
\end{figure}

We start by visualizing the differences between the space of features mapped between the two models. For a triplet network, this space is the sequence embeddings. Whilst, for the softmax model, we take the output of the 3072-dimensional last hidden layer. This layer is the concatenation between all convolutional layers and contains all the features used by the model. The two multi-dimensional vectors are reduced to 2D space using t-distributed stochastic neighbor embedding (tSNE). As mentioned in \citet{Alley2020}, the model is more accurate when the plasmid features are more separable in the latent (unobserved) space. We observe in figure \ref{fig:tsne} that the triplet network model has better-defined clusters.

It is also important to understand the robustness of the model, as small changes in plasmids can frequently occur. We perform random point mutations in the sequence from a lab and report the ranking of the correct lab generated by the model. Figure \ref{fig:effect-mutations} shows the mean and median of correct positions after 1000 runs for each perturbation. We found that virtually all runs with up to 100 mutations predicted the correct lab within the top 10 guesses. By increasing the number of mutations, the average of the runs becomes unstable, with the average of the positions being higher for all cases above 400 mutations. However, if we examine the median of the predicted positions, even with 1000 mutations, the median rank for the correct lab remains within the top-10. This indicates that our model is robust to most sequence perturbations, excluding cases where these mutations affect essential features for the model's prediction. 

We proceed to analyze all sequences to discover the importance of each plasmid feature to the model output. Unlike perturbation analysis in each sequence, here we use more recent methods that generate better insights in the interpretation of the model. Our method is based on integrated gradients \citep{integrated-grad}. The idea is to compute the gradient of the model's output relative to the embeddings token layer. This makes it possible to visualize the importance of each token for the model's prediction. After calculating the integrated gradient for all sequences, we obtain the token importance of each lab by averaging all sequences in that lab. The same process can be performed with all sequences to get the most seen tokens in the dataset.

\begin{figure*}[htp]
  \centering
\includegraphics[width=\linewidth]{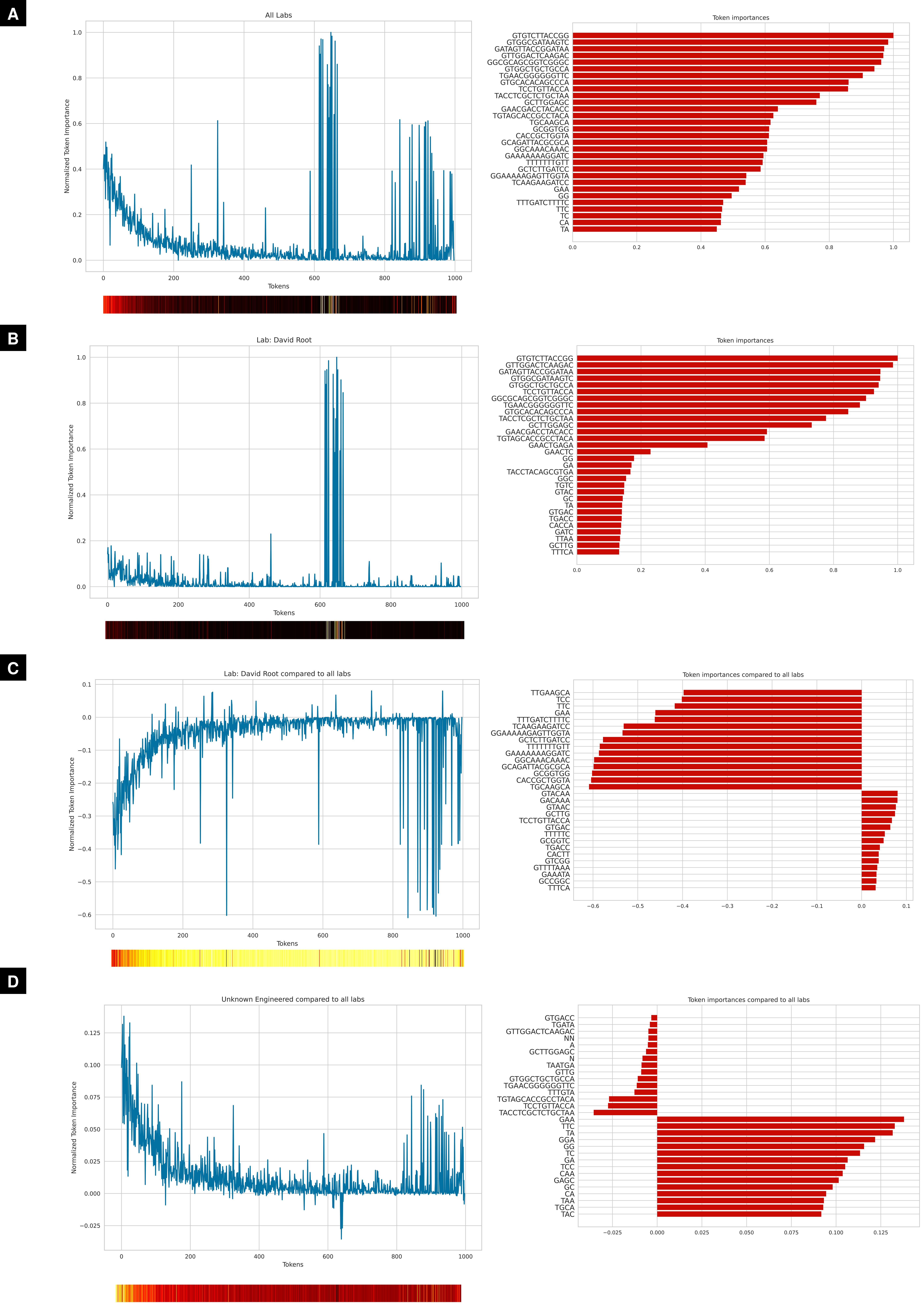}
  \caption{Interpreting plasmid features importance using integrated gradients.}
  \label{fig:model-interp}
\end{figure*}

\begin{figure}[htp]
  {\textbf{a} Normalized Token Importance (NTI) for all labs in the dataset obtained by averaging the token importance of all sequences and normalizing them between 0 and 1. On the right, the top 30 tokens for all data. \textbf{b} The normalized token importance for David Root's lab shows that it has similar tokens to all labs in the TOKEN ID range between 600 and 700, but in some other regions it differs a lot. \textbf{c} Difference between the token importance of David Root's lab and all labs. This graph highlights which tokens make a difference for this particular lab, whether considering the presence or absence of a token when compared to other labs. On the right, the tokens that should be observed when analyzing this lab.  \textbf{d} The token importance of the furthest lab to David Root's lab in the embedding space. The NTI values of the two laboratories are practically mirrored, indicating that these two laboratories have opposite characteristics.}
\end{figure}

As we can see from figure \ref{fig:model-interp} there are some tokens that appear to be shared by all labs. When generating the token importance of a lab, we can subtract it from the most seen tokens in the dataset to obtain a relative importance. This allows us to examine those tokens and those not to be expected for a particular lab. Furthermore, we can compare the token importance of one lab with the most distant lab from it in the embedded space. The graphs of token importance from the two labs are essentially mirrored, indicating that tokens are quite different in each case. Figure \ref{fig:model-interp} show these analyzes, where in the left-hand column, we plot Normalized Token Importance (NTI) as a function of the token. In the right hand column we highlight the sequences with the largest token importance. 

We further explored this model by looking at the token importance for David Root's lab (Figure \ref{fig:model-interp}B). From this analysis, we can see a cluster of sequences that are typical from this lab, allowing us to identify the potential design signatures. Furthermore, this lab is the furthest lab from the "unknown engineered" category. This class is a mixture of possible labs and so has poorly defined features. The fact that David Root's lab is the most different from this class suggests it has well-defined and perhaps has highly unique design choices. We note that for the "unknown engineered" class that the scale for the normalised token importance is shallow, and the colour gradient mostly red (see Figure \ref{fig:model-interp}D). This demonstrates that there is not a clear design choice or discriminating feature of this category, which is to be expected as it is a mixture of many possible labs (see methods). This analysis could be repeated for any of the labs in the dataset to identify key signatures or potential collaborations based on token proximity.

\begin{figure*}[htp]
  \centering
\includegraphics[width=\linewidth]{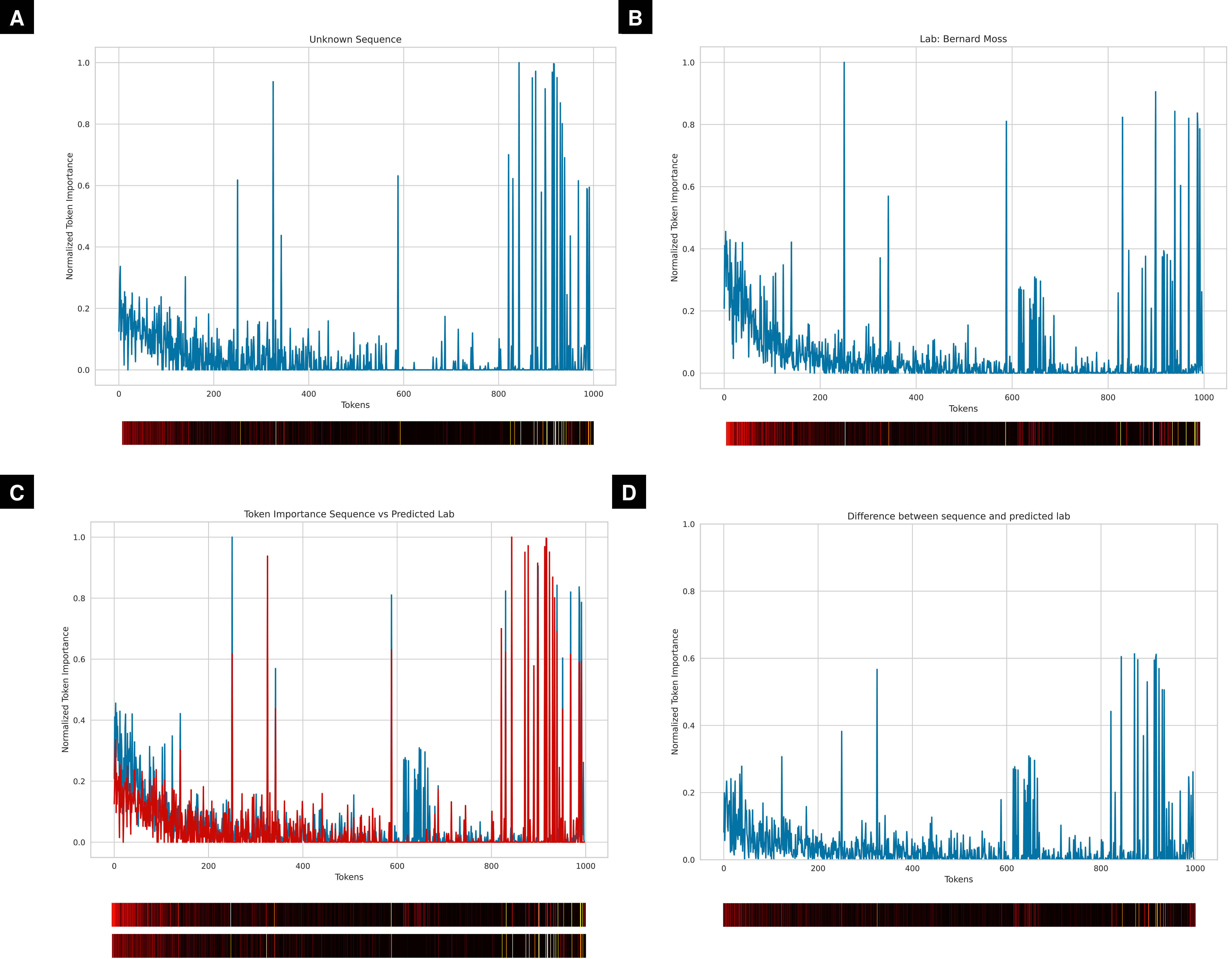}
  \caption{Plasmid features importance of unknown sequences. \textbf{a} Normalized Token Importance for an unknown sequence. This may help to investigate specific patterns. \textbf{b} The NTI for Bernard Moss's lab which was assigned the sequence author by our model. \textbf{c} Comparison between tokens highlighted for the sequence and important tokens from the predicted lab. The red line represents the sequence. \textbf{d} Plotting the difference between the NTI of the sequence and the NTI of the predicted laboratory, we can see that few tokens stand out in this sequence beyond the usual presented by the laboratory.}
  \label{fig:model-interp-2}
\end{figure*}

We next examined the use of integrated gradients for a single sequence. One of the major goals of GEA approaches, is to examine plasmids with unknown origin and be able to extract valuable sequence information, leading to correct assignment or further importation avenues to explore. Figure \ref{fig:model-interp-2} shows that with our approach we can obtain the importance of each token within an unknown sequence. When comparing the sequence token importance with the lab predicted by the model, we see a concordant behaviour in the plots demonstrating similarity in the highlighted features. This allows us to carefully examine the sequence features the model is using for prediction and hence allows secondary expert evaluation on the veracity of the prediction. 

\section{Discussion}

This manuscript presents a new state-of-the-art for genetic engineering attribution using convolutional neural networks and metric learning. We achieve 89.4\% top-10 lab prediction with a conventional classification model, by simply improving training details for better convergence and making use of ensemble learning by using parallel convolutions with different kernel sizes. Furthermore, we show that it is possible to treat the genetic engineering attribution problem as a metric learning problem, creating a vector space where genetic sequences with similar characteristics lie next to one another. Metric Learning is quite common in other areas such as recommendation systems \cite{recsys1, recsys2}. However, this is the first application to genetic engineering attribution. This methodology further improves the accuracy of our model, reaching 90.4\% top-10 accuracy, a 5.4\% improvement compared to the former state of the art and has several new advantages such as creating vector representation of labs, comparison and clustering of DNA sequences and labs-of-origin and the ability to examine design style and robustness to unseen labs. For example, a plasmid sequence might be too distant from known labs, resulting in low similarity values. Furthermore, we also have a particular embedding for unknown labs. If a new plasmid sequence is nearer to this embedding than to the known labs, it is possible that this sequence is from a currently unobserved lab. Meanwhile, a classifier model does not usually know how to handle such uncertainty. Typically, it spreads probabilities for each lab, summing up to $1.0$. Hence, any plasmid sequence is assigned to known labs, even if it is from a completely unknown lab.

Additionally, we demonstrated that following our methodology is also possible to perform few-shot learning. We achieve $58.1\%$ top-10 accuracy using only one sample, and with only 10\% of the sequences, we outperform the previous CNN approach without training a new model, simply by comparing embeddings vectors. This training methodology allows the possibility of identifying new laboratories with few samples and even just a single genetic sequence. Clearly, there is a tradeoff between sample quantity and model accuracy, but we believe that such methodology could be useful in extreme cases. Finally, these embeddings are also feature-rich, which means we can use them as input for other machine learning models, tackling other problems. For example, we are able to extract the defining signatures for labs and compare them to others using our approach. 

Although we present a new state-of-the-art classifier and a new training methodology with interpretability, we believe that there are possible improvements to our model. This could include the use of more advanced machine learning architectures, other pre-processing methods and new data augmentation techniques. This would lead to better convergence and algorithm training. We also believe that new techniques like Transformers \citep{Transformers::2017} and Graph Convolutional Networks \citep{Scarselli::2008} would be good candidates for this task, since patterns inside the sequence can be considered contextualized, for which Transformers generally show good performance \citep{Transformers::2017}. We hope our methodology and results encourage new architectures to tackle this problem.


\section{Method}

\subsection{Addgene dataset description and data splitting}

The addgene data was the same used by \citep{Alley2020}, and comprised all plasmids deposited in the Addgene repository up to July 27th 2018 – a total of 81,834 entries. For each plasmid, the dataset included a DNA sequence, along with metadata on growth strain, growth temperature, copy number, host species, bacterial resistance markers, and other selectable markers. Each of these categorical metadata fields was re-encoded as a series of one-hot feature groups:
\begin{itemize}
    \item Growth strain:  growth\_strain\_ccdb\_survival, growth\_strain\_dh10b, growth\_strain\_dh5alpha, growth\_strain\_neb\_stable, growth\_strain\_other, growth\_strain\_stbl3, growth\_strain\_top10, growth\_strain\_xl1\_blue
    \item Growth temperature: growth\_temp\_30, growth\_temp\_37, growth\_temp\_other
    \item Copy number: copy\_number\_high\_copy, copy\_number\_low\_copy, copy\_number\_unknown
    \item Host species:  species\_budding\_yeast, species\_fly, species\_human, species\_mouse,species\_mustard\_weed, species\_nematode, species\_other, species\_rat, species\_synthetic,species\_zebrafish
    \item Bacterial resistance:  bacterial\_resistance\_ampicillin, bacterial\_resistance\_chloramphenicol, bacterial\_resistance\_kanamycin, bacterial\_resistance\_other, bacterial\_resistance\_spectinomycin
    \item Other selectable markers:  selectable\_markers\_blasticidin, selectable\_markers\_his3, selectable\_markers\_hygromycin, selectable\_markers\_leu2, selectable\_markers\_neomycin, selectable\_markers\_other,selectable\_markers\_puromycin, selectable\_markers\_trp1, selectable\_markers\_ura3, selectable\_markers\_zeocin
\end{itemize}

In addition to the sequence and the above metadata fields, the raw dataset also contained unique sequence IDs, as well as separate IDs designating the origin lab. Both sequence and lab IDs were obfuscated through 1:1 replacement with random alphanumeric strings.
The number of plasmids deposited in the dataset by each lab was unbalanced, with many labs depositing one or a few sequences. To deal with this problem, \citet{Alley2020.08.22.262576} grouped labs with fewer than 10 data points into a single auxiliary category labelled “Unknown Engineered”. This reduced the number of categories from 3751 (the number of labs) to 1314 (1313 unique labs + Unknown Engineered).
In addition to issues with small labs, the dataset also contains “lineages” of plasmids. That is, sequences that were derived by modifying other sequences in the dataset. If unmitigated this introduces unintended correlations between the test and validation set. To overcome this, \citet{Alley2020.08.22.262576} inferred lineage networks among plasmids in the dataset, based on information in the complete Addgene database acknowledging sequence contributions from other entries. Lineages were identified by searching for connected components within the network of entry-to-entry acknowledgements in the Addgene database and we refer to \citet{Alley2020.08.22.262576} for more details. The data were partitioned into train, validation, and test sets, with the constraints that (i) every category have at least three data points in the test set, and (ii) all plasmids in a given lineage be assigned to a single dataset. Following the split, the training set contained 63,017 entries (77.0\%); the validation set contained 7,466 entries (9.1\%); and the test set contained 11,351 entries (13.9\%).

\subsection{Grouping sequences by their Levenshtein distance}\label{sec:grouping}

Genetic sequences from the same lab display large degrees of similarity. These sequences can make it easier to identify similar sequences in the training and validation sets. However, when training a machine learning algorithm, this may be perceived as data leakage between these sets \cite{leakage}, as the model does not need to learn to extract different features to identify such sequences.
To ameliorate this issue, we developed a more robust model by group sequence from each lab based on their Levenshtein distance  \cite{levdistance}. The Levenshtein formula used can be seen in equation \ref{Levensthein}.

\begin{equation}\label{Levensthein}
\operatorname{lev}_{a,b}(i,j)= \begin{cases} \max(i,j) & \text{ if } \min(i,j)=0, \\ \min \begin{cases} \operatorname{lev}_{a,b}(i-1,j) + 1 \\ \operatorname{lev}_{a,b}(i,j-1) + 1 \\ \operatorname{lev}_{a,b}(i-1,j-1) + 1_{(a_i \neq b_j)} \\ \end{cases} & \text{ otherwise.} \end{cases}
\end{equation}

After grouping, each laboratory has $N$ groups of sequences. We then split the dataset, ensuring that there will be sequences from the same group only in training or validation, which means they will never be present in both sets at the same time. This entire process was performed using python and the python-Levenshtein library (https://github.com/ztane/python-Levenshtein). As this is a costly algorithm and there are thousands of sequences to be grouped together, the entire process was performed on a machine with 128GB RAM and AMD EPYC 7401P 24-Core processor. This approach complements the lineage-based strategy, which also avoids data leakage.

\subsection{Training cross-validation}


We performed a 5-fold cross-validation strategy \cite{Hastie2009} for each experiment within the training set. To be precise, for each hyperparameter setting, we split the data into $k$ parts (we used $k=5$), using one of them to validate and the remaining to train, repeating this process $k$ times. After that, we evaluate each model, by taking the mean of the metrics for that experiment. This approach helps to avoid overfitting and improve generalisation. It also enables us to ensemble the $k$ models, to further improve generalisation.

\subsection{Byte pair encoding}

Similar to \citeauthor{Alley2020}, we use the Byte Pair Encoding algorithm \cite{gage1994} to process all sequences in the dataset, grouping subsequences into new tokens, increasing the vocabulary while reducing the sequences' size. The BPE algorithm first examines all the sequences in the dataset to learn how to perform the grouping. The trained algorithm is saved and used to transform the sequences during the convolutional neural network training. This last process is performed "online". This means that, while loading our batch of samples, we transform the batch into new small sequences. We converted our vocabulary size from 4 DNA bases into 1001 different tokens: 1000 tokens from the new vocabulary generated by the BPE algorithm plus 1 for an unknown token. The training and inference of the BPE algorithm were performed using the sentencepiece package (https://github.com/google/sentencepiece), and, as in the section \ref{sec:grouping}, we use python and the same machine to perform this operation. 

\subsection{Circular data augmentation}

Machine learning models and especially deep learning models are highly dependent on large amounts of data. One of the fundamental methods for adding variance to those models, increasing generalizability, and reducing overfitting \cite{Chicco2017} is data augmentation \cite{Mikolajczyk2018}. Generally, data augmentation performs transformations on the sample, considerably changing some characteristics. In this work, performing such transformations can be dangerous as it may end up modifying some essential parts to assign the sequence to a lab-of-origin. However, it is possible to take advantage of the fact that plasmids are circular and create a circular shift data augmentation process. This contrasts with a reverse complement augmentation one might use. During training, we show different versions of the same DNA sequence by shifting it circularly as shown in \ref{fig:arch}. This approach helps the model understand that the same pattern can happen at different positions within the sequence, increasing the generalizability of the training. Further, to decrease the model's complexity, we also limited the sequence to 1000 tokens.

We also perform Test Time Augmentation \cite{Moshkov2020} which helps to improve the model's prediction capability. To perform this analysis, during inference we run the model multiple times. Each time the model sees a shifted version of the sequence and makes a prediction of the same sample seen from different angles. We then take the average of the outputs (class probabilities for the classifier and embeddings for our proposed method). 

\subsection{CNN base architecture and training details}

Both types of models are composed of a Convolutional Neural Network with multiple kernels of different sizes, as proposed by \citeauthor{kim2014}. We use it to extract features from the sequence and concatenate them with the binary features provided in the dataset. The difference between the classification and triplet network models is on the final layers. The final base structure is composed of an embedding layer, several convolutional layers in parallel with different kernel sizes and custom dropout layer for regularization. The embedding layer has the shape of 1001x200, where 1001 is our vocabulary size and 200 the vector embedding dimension found empirically. Its purpose is to map each token into a 200-dimensional vector containing the features representation of that token \cite{Zou2018}. For the convolutional layers, we have a total of 12 layers in parallel, where the first layer has kernel size 1, the second has kernel size 2 and so on, until the last layer has kernel size 12. All convolutional layers are followed by a SeLU activation function \cite{selu} and a max pooling operation. We concatenate the features extracted by each of them, obtaining the final representation with different windowings of the sequence.  We also implemented a custom Dropout Layer. A standard Dropout Layer \cite{Srivastava2014} randomly masks out parts of a tensor to regularize the neural network. But if we did that on the embeddings before applying a similarity function, the output would be too unstable. So, we created a layer that randomly masks out the same parts of all the embeddings involved before applying the similarity function. We found this approach instrumental in regularizing our model.

The entire architecture was developed and trained using Python and Pytorch \cite{pytorch}. Although the training methodology is different between the two approaches, all the training details, such as optimizer, learning rate scheduler and regularization techniques remain the same. We use the Adam \cite{Adam} optimizer together with the One Cycle learning rate scheduler \cite{onecycle}. This scheduler was essential to achieve a better convergence in training and its settings were maximum learning rate of $1e^{-3}$ and cycle execution in $200$ epochs. To regularize our model and prevent overfitting, we used a weight decay of $1e^{-5}$ during training, and dropped $20\%$ of the embedded sequences using our custom dropout layer.

\subsection{Triplet network learning}

To generate the triplets, we use the labeled dataset to provide us with the anchor and positive. We then use a technique known as Hard Negative Mining \cite{hermans2017} to select the negative (an incorrect lab). This means that rather than choosing a random lab as a negative example, we choose the most challenging one given the current state of the embeddings. Thus, in our case it would be the nearest lab to our sequence in the latent space.

One of the most challenging parts of this work was the implementation of the algorithm to mine the negative examples during training efficiently. We could have used a library called PyTorch Metric Learning \cite{Musgrave2020}. They have Hard Negative Mining implemented per batch (it does not take the whole dataset into account while finding the negative) and implemented Cross-Batch Memory for Embedding Learning \cite{Wang2019}. However, this library only supports a single entity type. Furthermore, we also have easy access to the whole lab embeddings, since we use an embedding layer. So, the approach we re-implemented for our specific needs as described in the Algorithm \ref{alg:hard_negative_mining}. It is worth noting that we implemented it using tensors to make it as efficient as possible. Our source code provides a PyTorch implementation, and it should be straightforward to implement it in Tensorflow and other frameworks.

\begin{algorithm}[h]
  \SetAlgoNoEnd
  \DontPrintSemicolon
  \KwIn{$lab\_indices$ (B,), $anchor\_embeddings$ (B, L), $positive\_embeddings$ (B, E)}
  \KwOut{Batch of $negatives$ of shape (B, E)}
  $ all\_lab_indices \leftarrow $ indices of all labs repeated by B // (B, L) \;
  $ negative\_labs\_mask \leftarrow $ matrix of boolean values with $True$ for every lab // (B, L) \;
  $ negative\_labs\_mask[:, lab\_indices] \leftarrow False$ // (B, L) \;
  // At this point, we have a mask with True for the negative labs and False otherwise \;
  $ all\_negative\_lab\_indices \leftarrow $ $all\_labs[negative\_labs\_mask]$ reshaped // (B, L-1) \;
  $ all\_negative\_labs\_embeddings \leftarrow lab\_embeddings[all\_negative\_labs]$ L2 normalized // (B, L-1, E) \;
  $ anchor\_similarities \leftarrow $ dot product between $anchor\_embeddings$ and $all\_negative\_labs\_embeddings$ // (B, L-1) \;
  $ hardest\_negative\_lab_indices \leftarrow $ argmax of $anchor\_similarities$ // (B,) \;
  $ negative\_embeddings \leftarrow all\_negative\_labs\_embeddings[:, hardest\_negative\_lab\_indices]$ // (B, E) \;
  \Return{$negative\_embeddings$}
  
  \caption{Algorithm for Hard Negative Mining using tensors. The shape of each tensor is at the end of each line as a comment, being B batch size, E the embedding dimension, and L is the number of labs. It is worth noting that we L2 normalize all the embeddings.}\label{alg:hard_negative_mining}
\end{algorithm}

\subsection{Cosine similarity} \label{cosine}
We used the cosine similarity as the metric to measure how similar the vectors were in the embeddings. Mathematically, it measures the cosine of the angle between two vectors projected in a multi-dimensional space, resulting in a value inside the range of -1 and 1, where -1 indicates opposite vectors and 1 indicates equal vectors. Given two non-zeros vectors of embeddings, A and B, the cosine similarity is:

\begin{equation}
    {\text{similarity}}=\cos(\theta )={\mathbf {A} \cdot \mathbf {B}  \over \|\mathbf {A} \|\|\mathbf {B} \|}={\frac {\sum \limits _{i=1}^{n}{A_{i}B_{i}}}{{\sqrt {\sum \limits _{i=1}^{n}{A_{i}^{2}}}}{\sqrt {\sum \limits _{i=1}^{n}{B_{i}^{2}}}}}}
\end{equation}

\subsection{t-SNE and K-means}

Throughout this work, we have used $2$ techniques to analyze and visualize the results, the $t$-SNE algorithm, and the $K$-means cluster algorithm. T-Distributed Stochastic Neighbor Embedding \cite{tsne} is a dimensionality reduction technique that can reduce dimensions with non-linear relationships. It is particularly well suited for the visualization of high-dimensional complex real-world datasets. Making use of this, we are able to reduce the embeddings generated by the triplet network and visualize them in 2D space. K-means \cite{kmeans} is a clustering algorithm that attempts to organize the data into K clusters. The objective of the algorithm is to group similar data together by their Euclidean distance to $K$ centroids, where K is a value chosen by the user, while keeping all centroids distant from each other. Each sample will be linked (or allocated/assigned) to the cluster with the nearest centroid. 

\subsection{Interpreting the model}

To visualize the mapped features of both models, we took different approaches since they are different models. For the triplet network model, we first inferred all sequences from the validation set and got their embeddings. These embeddings were 200-dimensional and they were reduced to 2 dimensions using tSNE in scikit-learn project \cite{scikit-learn} with default parameters. For the classification model, we extracted the activations of the last hidden layer that maps features from all convolutional layers, before concatenating those features with the extra inputs (sequence metadata) and passing through the last layer which outputs logits. These hidden features were 3072-dimensional, reduced to 2 dimensions in the same way as embeddings. The visualization was done using matplotlib \cite{matplotlib} and we colored each point by the corresponding lab.

To analyze the influence of perturbations on model prediction, we took a specific plasmid from the validation set and randomly generated perturbation in its sequence. As the process is random, we perform 100 experiments for each number of mutations, ranging from 1 to 1000. To make these mutations, we use a random integer function to select the specific position to be changed and a random choice operation to choose one of the 5 possible bases in the sequence (N, G, C, T, A). For each of the 100,000 runs, we took the position of the correct laboratory in the model's prediction ranking. The figure \ref{fig:effect-mutations} presents the mean and median of those positions.

To find out which tokens are most important within a sequence, we decided to use a similar methodology to integrated gradients \cite{integrated-grad}. Integrated gradient is an interpretability technique for deep neural networks which finds the input features that contribute the most to the model prediction. We started by computing gradients between model predictions with respect to the sequence embedding layer, getting a matrix of gradients in the shape of 1001x200 (number of tokens x embedding dimension). Each gradient measures the relationship between the embedding weight and the output. We then calculate the absolute value element-wise and sum them up in the second axis, generating a final vector of 1001 positions containing the summed importance of each token.  

\begin{equation}
    {TokenImportance_{i}(x) = \sum_{k=1}^{200}\|\frac{\partial F(x)}{\partial x_{ik} }\| \times \frac{1}{200}}
\end{equation}

where: 

$F()$ = model's prediction function

$\frac{\partial F}{\partial x_{ik} }$  = gradient of model F's prediction function relative to each embedding feature $x_{ik}$

$i$ = number of token

$k$ = embedidng dimension position 

After generating the token importance of each sequence in the validation set, we took the token importance of each laboratory by averaging the token importance of all sequences in that laboratory. The visualization was done using matplotlib, and to better present the figure, the token importance values were normalized between 0 and 1 (NTI). To compare the NTI of a specific lab with the further lab from it, we compute the cosine similarity between the analyzed lab embedding and all other lab embeddings by performing a dot product. The lowest value indicates the least similar laboratory. 

\begin{acks}
To Amalgam and XNV for providing us the necessary infrastructure and financial support. OMC acknowledges funding from a Todd-Bird Junior Research Fellowship from New College, Oxford, as well as Open Philanthropy. 
\end{acks}

\bibliographystyle{ACM-Reference-Format}
\bibliography{bib/references.bib}


\begin{thebibliography}{48}


\ifx \showCODEN    \undefined \def \showCODEN     #1{\unskip}     \fi
\ifx \showDOI      \undefined \def \showDOI       #1{#1}\fi
\ifx \showISBNx    \undefined \def \showISBNx     #1{\unskip}     \fi
\ifx \showISBNxiii \undefined \def \showISBNxiii  #1{\unskip}     \fi
\ifx \showISSN     \undefined \def \showISSN      #1{\unskip}     \fi
\ifx \showLCCN     \undefined \def \showLCCN      #1{\unskip}     \fi
\ifx \shownote     \undefined \def \shownote      #1{#1}          \fi
\ifx \showarticletitle \undefined \def \showarticletitle #1{#1}   \fi
\ifx \showURL      \undefined \def \showURL       {\relax}        \fi
\providecommand\bibfield[2]{#2}
\providecommand\bibinfo[2]{#2}
\providecommand\natexlab[1]{#1}
\providecommand\showeprint[2][]{arXiv:#2}

\bibitem[\protect\citeauthoryear{Alley, Turpin, Liu, Kulp-McDowall, Swett,
  Edison, Von~Stetina, Church, and Esvelt}{Alley et~al\mbox{.}}{2020a}]%
        {Alley2020.08.22.262576}
\bibfield{author}{\bibinfo{person}{Ethan~C. Alley}, \bibinfo{person}{Miles
  Turpin}, \bibinfo{person}{Andrew~Bo Liu}, \bibinfo{person}{Taylor
  Kulp-McDowall}, \bibinfo{person}{Jacob Swett}, \bibinfo{person}{Rey Edison},
  \bibinfo{person}{Stephen~E. Von~Stetina}, \bibinfo{person}{George~M. Church},
  {and} \bibinfo{person}{Kevin~M. Esvelt}.} \bibinfo{year}{2020}\natexlab{a}.
\newblock \showarticletitle{Attribution of genetic engineering: A practical and
  accurate machine-learning toolkit for biosecurity}.
\newblock \bibinfo{journal}{\emph{bioRxiv}} (\bibinfo{year}{2020}).
\newblock
\urldef\tempurl%
\url{https://doi.org/10.1101/2020.08.22.262576}
\showDOI{\tempurl}
\showeprint{https://www.biorxiv.org/content/early/2020/08/22/2020.08.22.262576.full.pdf}


\bibitem[\protect\citeauthoryear{Alley, Turpin, Liu, Kulp-McDowall, Swett,
  Edison, Von~Stetina, Church, and Esvelt}{Alley et~al\mbox{.}}{2020b}]%
        {Alley2020}
\bibfield{author}{\bibinfo{person}{Ethan~C. Alley}, \bibinfo{person}{Miles
  Turpin}, \bibinfo{person}{Andrew~Bo Liu}, \bibinfo{person}{Taylor
  Kulp-McDowall}, \bibinfo{person}{Jacob Swett}, \bibinfo{person}{Rey Edison},
  \bibinfo{person}{Stephen~E. Von~Stetina}, \bibinfo{person}{George~M. Church},
  {and} \bibinfo{person}{Kevin~M. Esvelt}.} \bibinfo{year}{2020}\natexlab{b}.
\newblock \showarticletitle{A machine learning toolkit for genetic engineering
  attribution to facilitate biosecurity}.
\newblock \bibinfo{journal}{\emph{Nature Communications}} \bibinfo{volume}{11},
  \bibinfo{number}{1} (\bibinfo{date}{08 Dec} \bibinfo{year}{2020}),
  \bibinfo{pages}{6293}.
\newblock
\showISSN{2041-1723}
\urldef\tempurl%
\url{https://doi.org/10.1038/s41467-020-19612-0}
\showDOI{\tempurl}


\bibitem[\protect\citeauthoryear{Altschul, Gish, Miller, Myers, and
  Lipman}{Altschul et~al\mbox{.}}{1990}]%
        {Altschul1990}
\bibfield{author}{\bibinfo{person}{Stephen~F. Altschul},
  \bibinfo{person}{Warren Gish}, \bibinfo{person}{Webb Miller},
  \bibinfo{person}{Eugene~W. Myers}, {and} \bibinfo{person}{David~J. Lipman}.}
  \bibinfo{year}{1990}\natexlab{}.
\newblock \showarticletitle{Basic local alignment search tool}.
\newblock \bibinfo{journal}{\emph{Journal of Molecular Biology}}
  \bibinfo{volume}{215}, \bibinfo{number}{3} (\bibinfo{date}{Oct.}
  \bibinfo{year}{1990}), \bibinfo{pages}{403--410}.
\newblock
\urldef\tempurl%
\url{https://doi.org/10.1016/s0022-2836(05)80360-2}
\showDOI{\tempurl}


\bibitem[\protect\citeauthoryear{Arthur and Vassilvitskii}{Arthur and
  Vassilvitskii}{2007}]%
        {kmeans}
\bibfield{author}{\bibinfo{person}{David Arthur} {and} \bibinfo{person}{Sergei
  Vassilvitskii}.} \bibinfo{year}{2007}\natexlab{}.
\newblock \showarticletitle{K-Means++: The Advantages of Careful Seeding}. In
  \bibinfo{booktitle}{\emph{Proceedings of the Eighteenth Annual ACM-SIAM
  Symposium on Discrete Algorithms}} (New Orleans, Louisiana)
  \emph{(\bibinfo{series}{SODA '07})}. \bibinfo{publisher}{Society for
  Industrial and Applied Mathematics}, \bibinfo{address}{USA},
  \bibinfo{pages}{1027–1035}.
\newblock
\showISBNx{9780898716245}


\bibitem[\protect\citeauthoryear{Berger, Waterman, and Yu}{Berger
  et~al\mbox{.}}{2020}]%
        {levdistance}
\bibfield{author}{\bibinfo{person}{Bonnie Berger}, \bibinfo{person}{Michael
  Waterman}, {and} \bibinfo{person}{Yun Yu}.} \bibinfo{year}{2020}\natexlab{}.
\newblock \showarticletitle{Levenshtein Distance, Sequence Comparison and
  Biological Database Search}.
\newblock \bibinfo{journal}{\emph{IEEE Transactions on Information Theory}}
  \bibinfo{volume}{PP} (\bibinfo{date}{05} \bibinfo{year}{2020}),
  \bibinfo{pages}{1--1}.
\newblock
\urldef\tempurl%
\url{https://doi.org/10.1109/TIT.2020.2996543}
\showDOI{\tempurl}


\bibitem[\protect\citeauthoryear{Chakraborty, Tomsett, Raghavendra, Harborne,
  Alzantot, Cerutti, Srivastava, Preece, Julier, Rao, Kelley, Braines, Sensoy,
  Willis, and Gurram}{Chakraborty et~al\mbox{.}}{2017}]%
        {Chakraborty2017InterpretabilityOD}
\bibfield{author}{\bibinfo{person}{Supriyo Chakraborty},
  \bibinfo{person}{Richard~J. Tomsett}, \bibinfo{person}{Ramya Raghavendra},
  \bibinfo{person}{Daniel Harborne}, \bibinfo{person}{Moustafa Alzantot},
  \bibinfo{person}{F. Cerutti}, \bibinfo{person}{Mani~B. Srivastava},
  \bibinfo{person}{Alun~David Preece}, \bibinfo{person}{Simon~J. Julier},
  \bibinfo{person}{Raghuveer~M. Rao}, \bibinfo{person}{Troy~D. Kelley},
  \bibinfo{person}{Dave Braines}, \bibinfo{person}{M. Sensoy},
  \bibinfo{person}{Chris~J. Willis}, {and} \bibinfo{person}{Prudhvi~K.
  Gurram}.} \bibinfo{year}{2017}\natexlab{}.
\newblock \showarticletitle{Interpretability of deep learning models: A survey
  of results}.
\newblock \bibinfo{journal}{\emph{2017 IEEE SmartWorld, Ubiquitous Intelligence
  \& Computing, Advanced \& Trusted Computed, Scalable Computing \&
  Communications, Cloud \& Big Data Computing, Internet of People and Smart
  City Innovation (SmartWorld/SCALCOM/UIC/ATC/CBDCom/IOP/SCI)}}
  (\bibinfo{year}{2017}), \bibinfo{pages}{1--6}.
\newblock


\bibitem[\protect\citeauthoryear{Chicco}{Chicco}{2017}]%
        {Chicco2017}
\bibfield{author}{\bibinfo{person}{Davide Chicco}.}
  \bibinfo{year}{2017}\natexlab{}.
\newblock \showarticletitle{Ten quick tips for machine learning in
  computational biology}.
\newblock \bibinfo{journal}{\emph{{BioData} Mining}} \bibinfo{volume}{10},
  \bibinfo{number}{1} (\bibinfo{date}{Dec.} \bibinfo{year}{2017}).
\newblock
\urldef\tempurl%
\url{https://doi.org/10.1186/s13040-017-0155-3}
\showDOI{\tempurl}


\bibitem[\protect\citeauthoryear{Doshi-Velez and Kim}{Doshi-Velez and
  Kim}{2017}]%
        {doshivelez2017rigorous}
\bibfield{author}{\bibinfo{person}{Finale Doshi-Velez} {and}
  \bibinfo{person}{Been Kim}.} \bibinfo{year}{2017}\natexlab{}.
\newblock \bibinfo{title}{Towards A Rigorous Science of Interpretable Machine
  Learning}.
\newblock
\newblock
\showeprint[arxiv]{1702.08608}~[stat.ML]


\bibitem[\protect\citeauthoryear{Fei-Fei, Fergus, and Perona}{Fei-Fei
  et~al\mbox{.}}{2006}]%
        {LiFeiFei2006}
\bibfield{author}{\bibinfo{person}{Li Fei-Fei}, \bibinfo{person}{R. Fergus},
  {and} \bibinfo{person}{P. Perona}.} \bibinfo{year}{2006}\natexlab{}.
\newblock \showarticletitle{One-shot learning of object categories}.
\newblock \bibinfo{journal}{\emph{{IEEE} Transactions on Pattern Analysis and
  Machine Intelligence}} \bibinfo{volume}{28}, \bibinfo{number}{4}
  (\bibinfo{date}{April} \bibinfo{year}{2006}), \bibinfo{pages}{594--611}.
\newblock
\urldef\tempurl%
\url{https://doi.org/10.1109/tpami.2006.79}
\showDOI{\tempurl}


\bibitem[\protect\citeauthoryear{Fink}{Fink}{2005}]%
        {NIPS2004_ef1e491a}
\bibfield{author}{\bibinfo{person}{Michael Fink}.}
  \bibinfo{year}{2005}\natexlab{}.
\newblock \showarticletitle{Object Classification from a Single Example
  Utilizing Class Relevance Metrics}. In \bibinfo{booktitle}{\emph{Advances in
  Neural Information Processing Systems}},
  \bibfield{editor}{\bibinfo{person}{L.~Saul}, \bibinfo{person}{Y.~Weiss},
  {and} \bibinfo{person}{L.~Bottou}} (Eds.), Vol.~\bibinfo{volume}{17}.
  \bibinfo{publisher}{MIT Press}.
\newblock
\urldef\tempurl%
\url{https://proceedings.neurips.cc/paper/2004/file/ef1e491a766ce3127556063d49bc2f98-Paper.pdf}
\showURL{%
\tempurl}


\bibitem[\protect\citeauthoryear{Gage}{Gage}{1994}]%
        {gage1994}
\bibfield{author}{\bibinfo{person}{Philip Gage}.}
  \bibinfo{year}{1994}\natexlab{}.
\newblock \showarticletitle{A New Algorithm for Data Compression}.
\newblock \bibinfo{journal}{\emph{C Users J.}} \bibinfo{volume}{12},
  \bibinfo{number}{2} (\bibinfo{date}{Feb.} \bibinfo{year}{1994}),
  \bibinfo{pages}{23–38}.
\newblock
\showISSN{0898-9788}


\bibitem[\protect\citeauthoryear{Hastie, Tibshirani, and Friedman}{Hastie
  et~al\mbox{.}}{2009}]%
        {Hastie2009}
\bibfield{author}{\bibinfo{person}{Trevor Hastie}, \bibinfo{person}{Robert
  Tibshirani}, {and} \bibinfo{person}{Jerome Friedman}.}
  \bibinfo{year}{2009}\natexlab{}.
\newblock \bibinfo{booktitle}{\emph{The elements of statistical learning: data
  mining, inference and prediction} (\bibinfo{edition}{2} ed.)}.
\newblock \bibinfo{publisher}{Springer}.
\newblock
\urldef\tempurl%
\url{http://www-stat.stanford.edu/~tibs/ElemStatLearn/}
\showURL{%
\tempurl}


\bibitem[\protect\citeauthoryear{Hermans, Beyer, and Leibe}{Hermans
  et~al\mbox{.}}{2017}]%
        {hermans2017}
\bibfield{author}{\bibinfo{person}{Alexander Hermans}, \bibinfo{person}{Lucas
  Beyer}, {and} \bibinfo{person}{Bastian Leibe}.}
  \bibinfo{year}{2017}\natexlab{}.
\newblock \bibinfo{title}{In Defense of the Triplet Loss for Person
  Re-Identification}.
\newblock
\newblock
\showeprint[arxiv]{1703.07737}~[cs.CV]


\bibitem[\protect\citeauthoryear{Hinton and Roweis}{Hinton and Roweis}{2002}]%
        {hinton2002}
\bibfield{author}{\bibinfo{person}{Geoffrey Hinton} {and} \bibinfo{person}{Sam
  Roweis}.} \bibinfo{year}{2002}\natexlab{}.
\newblock \showarticletitle{Stochastic Neighbor Embedding}. In
  \bibinfo{booktitle}{\emph{Proceedings of the 15th International Conference on
  Neural Information Processing Systems}} \emph{(\bibinfo{series}{NIPS'02})}.
  \bibinfo{publisher}{MIT Press}, \bibinfo{address}{Cambridge, MA, USA},
  \bibinfo{pages}{857–864}.
\newblock


\bibitem[\protect\citeauthoryear{Hoffer and Ailon}{Hoffer and Ailon}{2018}]%
        {hoffer2018deep}
\bibfield{author}{\bibinfo{person}{Elad Hoffer} {and} \bibinfo{person}{Nir
  Ailon}.} \bibinfo{year}{2018}\natexlab{}.
\newblock \bibinfo{title}{Deep metric learning using Triplet network}.
\newblock
\newblock
\showeprint[arxiv]{1412.6622}~[cs.LG]


\bibitem[\protect\citeauthoryear{Hsieh, Yang, Cui, Lin, Belongie, and
  Estrin}{Hsieh et~al\mbox{.}}{2017}]%
        {recsys1}
\bibfield{author}{\bibinfo{person}{Cheng-Kang Hsieh}, \bibinfo{person}{Longqi
  Yang}, \bibinfo{person}{Yin Cui}, \bibinfo{person}{Tsung-Yi Lin},
  \bibinfo{person}{Serge Belongie}, {and} \bibinfo{person}{Deborah Estrin}.}
  \bibinfo{year}{2017}\natexlab{}.
\newblock \showarticletitle{Collaborative Metric Learning}.
  \bibinfo{publisher}{International World Wide Web Conferences Steering
  Committee}.
\newblock
\urldef\tempurl%
\url{https://doi.org/10.1145/3038912.3052639}
\showDOI{\tempurl}


\bibitem[\protect\citeauthoryear{Hunter}{Hunter}{2007}]%
        {matplotlib}
\bibfield{author}{\bibinfo{person}{J.~D. Hunter}.}
  \bibinfo{year}{2007}\natexlab{}.
\newblock \showarticletitle{Matplotlib: A 2D graphics environment}.
\newblock \bibinfo{journal}{\emph{Computing in Science \& Engineering}}
  \bibinfo{volume}{9}, \bibinfo{number}{3} (\bibinfo{year}{2007}),
  \bibinfo{pages}{90--95}.
\newblock
\urldef\tempurl%
\url{https://doi.org/10.1109/MCSE.2007.55}
\showDOI{\tempurl}


\bibitem[\protect\citeauthoryear{Kamens}{Kamens}{2014}]%
        {Addgene}
\bibfield{author}{\bibinfo{person}{Joanne Kamens}.}
  \bibinfo{year}{2014}\natexlab{}.
\newblock \showarticletitle{The Addgene repository: an international nonprofit
  plasmid and data resource}.
\newblock  \bibinfo{volume}{43}, \bibinfo{number}{D1} (\bibinfo{date}{Nov.}
  \bibinfo{year}{2014}), \bibinfo{pages}{D1152--D1157}.
\newblock
\urldef\tempurl%
\url{https://doi.org/10.1093/nar/gku893}
\showDOI{\tempurl}


\bibitem[\protect\citeauthoryear{Karim, Beyan, Zappa, Costa, Rebholz-Schuhmann,
  Cochez, and Decker}{Karim et~al\mbox{.}}{2020}]%
        {clustering}
\bibfield{author}{\bibinfo{person}{Md~Rezaul Karim}, \bibinfo{person}{Oya
  Beyan}, \bibinfo{person}{Achille Zappa}, \bibinfo{person}{Ivan~G Costa},
  \bibinfo{person}{Dietrich Rebholz-Schuhmann}, \bibinfo{person}{Michael
  Cochez}, {and} \bibinfo{person}{Stefan Decker}.}
  \bibinfo{year}{2020}\natexlab{}.
\newblock \showarticletitle{Deep learning-based clustering approaches for
  bioinformatics}.
\newblock  \bibinfo{volume}{22}, \bibinfo{number}{1} (\bibinfo{date}{Feb.}
  \bibinfo{year}{2020}), \bibinfo{pages}{393--415}.
\newblock
\urldef\tempurl%
\url{https://doi.org/10.1093/bib/bbz170}
\showDOI{\tempurl}


\bibitem[\protect\citeauthoryear{Kaufman, Rosset, and Perlich}{Kaufman
  et~al\mbox{.}}{2011}]%
        {leakage}
\bibfield{author}{\bibinfo{person}{Shachar Kaufman}, \bibinfo{person}{Saharon
  Rosset}, {and} \bibinfo{person}{Claudia Perlich}.}
  \bibinfo{year}{2011}\natexlab{}.
\newblock \showarticletitle{Leakage in Data Mining: Formulation, Detection, and
  Avoidance}. In \bibinfo{booktitle}{\emph{Proceedings of the 17th ACM SIGKDD
  International Conference on Knowledge Discovery and Data Mining}} (San Diego,
  California, USA) \emph{(\bibinfo{series}{KDD '11})}.
  \bibinfo{publisher}{Association for Computing Machinery},
  \bibinfo{address}{New York, NY, USA}, \bibinfo{pages}{556–563}.
\newblock
\showISBNx{9781450308137}
\urldef\tempurl%
\url{https://doi.org/10.1145/2020408.2020496}
\showDOI{\tempurl}


\bibitem[\protect\citeauthoryear{Kim}{Kim}{2014}]%
        {kim2014}
\bibfield{author}{\bibinfo{person}{Yoon Kim}.} \bibinfo{year}{2014}\natexlab{}.
\newblock \bibinfo{title}{Convolutional Neural Networks for Sentence
  Classification}.
\newblock
\newblock
\showeprint[arxiv]{1408.5882}~[cs.CL]


\bibitem[\protect\citeauthoryear{Kingma and Ba}{Kingma and Ba}{2014}]%
        {Adam}
\bibfield{author}{\bibinfo{person}{Diederik~P. Kingma} {and}
  \bibinfo{person}{Jimmy Ba}.} \bibinfo{year}{2014}\natexlab{}.
\newblock \bibinfo{title}{Adam: A Method for Stochastic Optimization}.
\newblock
\newblock
\urldef\tempurl%
\url{http://arxiv.org/abs/1412.6980}
\showURL{%
\tempurl}
\newblock
\shownote{cite arxiv:1412.6980Comment: Published as a conference paper at the
  3rd International Conference for Learning Representations, San Diego, 2015}.


\bibitem[\protect\citeauthoryear{Klambauer, Unterthiner, Mayr, and
  Hochreiter}{Klambauer et~al\mbox{.}}{2017}]%
        {selu}
\bibfield{author}{\bibinfo{person}{G{\"{u}}nter Klambauer},
  \bibinfo{person}{Thomas Unterthiner}, \bibinfo{person}{Andreas Mayr}, {and}
  \bibinfo{person}{Sepp Hochreiter}.} \bibinfo{year}{2017}\natexlab{}.
\newblock \showarticletitle{Self-Normalizing Neural Networks}.
\newblock \bibinfo{journal}{\emph{CoRR}}  \bibinfo{volume}{abs/1706.02515}
  (\bibinfo{year}{2017}).
\newblock
\showeprint[arXiv]{1706.02515}
\urldef\tempurl%
\url{http://arxiv.org/abs/1706.02515}
\showURL{%
\tempurl}


\bibitem[\protect\citeauthoryear{Koch, Zemel, and Salakhutdinov}{Koch
  et~al\mbox{.}}{2015}]%
        {one-shot}
\bibfield{author}{\bibinfo{person}{Gregory Koch}, \bibinfo{person}{Richard
  Zemel}, {and} \bibinfo{person}{Ruslan Salakhutdinov}.}
  \bibinfo{year}{2015}\natexlab{}.
\newblock \showarticletitle{Siamese Neural Networks for One-shot Image
  Recognition}.
\newblock


\bibitem[\protect\citeauthoryear{Kulis}{Kulis}{2013}]%
        {kulis2013}
\bibfield{author}{\bibinfo{person}{Brian Kulis}.}
  \bibinfo{year}{2013}\natexlab{}.
\newblock \showarticletitle{Metric Learning: A Survey}.
\newblock \bibinfo{journal}{\emph{Foundations and Trends® in Machine
  Learning}} \bibinfo{volume}{5}, \bibinfo{number}{4} (\bibinfo{year}{2013}),
  \bibinfo{pages}{287--364}.
\newblock
\showISSN{1935-8237}
\urldef\tempurl%
\url{https://doi.org/10.1561/2200000019}
\showDOI{\tempurl}


\bibitem[\protect\citeauthoryear{Lipton}{Lipton}{2015}]%
        {rnn1}
\bibfield{author}{\bibinfo{person}{Zachary~Chase Lipton}.}
  \bibinfo{year}{2015}\natexlab{}.
\newblock \showarticletitle{A Critical Review of Recurrent Neural Networks for
  Sequence Learning}.
\newblock \bibinfo{journal}{\emph{CoRR}}  \bibinfo{volume}{abs/1506.00019}
  (\bibinfo{year}{2015}).
\newblock
\showeprint[arxiv]{1506.00019}
\urldef\tempurl%
\url{http://arxiv.org/abs/1506.00019}
\showURL{%
\tempurl}


\bibitem[\protect\citeauthoryear{Mikolajczyk and Grochowski}{Mikolajczyk and
  Grochowski}{2018}]%
        {Mikolajczyk2018}
\bibfield{author}{\bibinfo{person}{Agnieszka Mikolajczyk} {and}
  \bibinfo{person}{Michal Grochowski}.} \bibinfo{year}{2018}\natexlab{}.
\newblock \showarticletitle{Data augmentation for improving deep learning in
  image classification problem}. In \bibinfo{booktitle}{\emph{2018
  International Interdisciplinary {PhD} Workshop ({IIPhDW})}}.
  \bibinfo{publisher}{{IEEE}}.
\newblock
\urldef\tempurl%
\url{https://doi.org/10.1109/iiphdw.2018.8388338}
\showDOI{\tempurl}


\bibitem[\protect\citeauthoryear{Mikolov, Sutskever, Chen, Corrado, and
  Dean}{Mikolov et~al\mbox{.}}{2013}]%
        {wordembedding}
\bibfield{author}{\bibinfo{person}{Tomas Mikolov}, \bibinfo{person}{Ilya
  Sutskever}, \bibinfo{person}{Kai Chen}, \bibinfo{person}{Greg~S Corrado},
  {and} \bibinfo{person}{Jeff Dean}.} \bibinfo{year}{2013}\natexlab{}.
\newblock \showarticletitle{Distributed Representations of Words and Phrases
  and their Compositionality}. In \bibinfo{booktitle}{\emph{Advances in Neural
  Information Processing Systems}}, \bibfield{editor}{\bibinfo{person}{C.~J.~C.
  Burges}, \bibinfo{person}{L.~Bottou}, \bibinfo{person}{M.~Welling},
  \bibinfo{person}{Z.~Ghahramani}, {and} \bibinfo{person}{K.~Q. Weinberger}}
  (Eds.), Vol.~\bibinfo{volume}{26}. \bibinfo{publisher}{Curran Associates,
  Inc.}
\newblock
\urldef\tempurl%
\url{https://proceedings.neurips.cc/paper/2013/file/9aa42b31882ec039965f3c4923ce901b-Paper.pdf}
\showURL{%
\tempurl}


\bibitem[\protect\citeauthoryear{Moshkov, Mathe, Kertesz-Farkas, Hollandi, and
  Horvath}{Moshkov et~al\mbox{.}}{2020}]%
        {Moshkov2020}
\bibfield{author}{\bibinfo{person}{Nikita Moshkov}, \bibinfo{person}{Botond
  Mathe}, \bibinfo{person}{Attila Kertesz-Farkas}, \bibinfo{person}{Reka
  Hollandi}, {and} \bibinfo{person}{Peter Horvath}.}
  \bibinfo{year}{2020}\natexlab{}.
\newblock \showarticletitle{Test-time augmentation for deep learning-based cell
  segmentation on microscopy images}.
\newblock \bibinfo{journal}{\emph{Scientific Reports}} \bibinfo{volume}{10},
  \bibinfo{number}{1} (\bibinfo{date}{March} \bibinfo{year}{2020}).
\newblock
\urldef\tempurl%
\url{https://doi.org/10.1038/s41598-020-61808-3}
\showDOI{\tempurl}


\bibitem[\protect\citeauthoryear{Musgrave, Belongie, and Lim}{Musgrave
  et~al\mbox{.}}{2020}]%
        {Musgrave2020}
\bibfield{author}{\bibinfo{person}{Kevin Musgrave}, \bibinfo{person}{Serge
  Belongie}, {and} \bibinfo{person}{Ser-Nam Lim}.}
  \bibinfo{year}{2020}\natexlab{}.
\newblock \bibinfo{title}{PyTorch Metric Learning}.
\newblock
\newblock
\showeprint[arxiv]{2008.09164}~[cs.CV]


\bibitem[\protect\citeauthoryear{Nielsen and Voigt}{Nielsen and Voigt}{2018}]%
        {Nielsen2018}
\bibfield{author}{\bibinfo{person}{Alec A.~K. Nielsen} {and}
  \bibinfo{person}{Christopher~A. Voigt}.} \bibinfo{year}{2018}\natexlab{}.
\newblock \showarticletitle{Deep learning to predict the lab-of-origin of
  engineered DNA}.
\newblock \bibinfo{journal}{\emph{Nature Communications}} \bibinfo{volume}{9},
  \bibinfo{number}{1} (\bibinfo{date}{07 Aug} \bibinfo{year}{2018}),
  \bibinfo{pages}{3135}.
\newblock
\showISSN{2041-1723}
\urldef\tempurl%
\url{https://doi.org/10.1038/s41467-018-05378-z}
\showDOI{\tempurl}


\bibitem[\protect\citeauthoryear{Omran, Engelbrecht, and Salman}{Omran
  et~al\mbox{.}}{2007}]%
        {clustering3}
\bibfield{author}{\bibinfo{person}{Mahamed Omran}, \bibinfo{person}{Andries
  Engelbrecht}, {and} \bibinfo{person}{Ayed Salman}.}
  \bibinfo{year}{2007}\natexlab{}.
\newblock \showarticletitle{An overview of clustering methods}.
\newblock \bibinfo{journal}{\emph{Intell. Data Anal.}}  \bibinfo{volume}{11}
  (\bibinfo{date}{11} \bibinfo{year}{2007}), \bibinfo{pages}{583--605}.
\newblock
\urldef\tempurl%
\url{https://doi.org/10.3233/IDA-2007-11602}
\showDOI{\tempurl}


\bibitem[\protect\citeauthoryear{Paszke, Gross, Massa, Lerer, Bradbury, Chanan,
  Killeen, Lin, Gimelshein, Antiga, Desmaison, Kopf, Yang, DeVito, Raison,
  Tejani, Chilamkurthy, Steiner, Fang, Bai, and Chintala}{Paszke
  et~al\mbox{.}}{2019}]%
        {pytorch}
\bibfield{author}{\bibinfo{person}{Adam Paszke}, \bibinfo{person}{Sam Gross},
  \bibinfo{person}{Francisco Massa}, \bibinfo{person}{Adam Lerer},
  \bibinfo{person}{James Bradbury}, \bibinfo{person}{Gregory Chanan},
  \bibinfo{person}{Trevor Killeen}, \bibinfo{person}{Zeming Lin},
  \bibinfo{person}{Natalia Gimelshein}, \bibinfo{person}{Luca Antiga},
  \bibinfo{person}{Alban Desmaison}, \bibinfo{person}{Andreas Kopf},
  \bibinfo{person}{Edward Yang}, \bibinfo{person}{Zachary DeVito},
  \bibinfo{person}{Martin Raison}, \bibinfo{person}{Alykhan Tejani},
  \bibinfo{person}{Sasank Chilamkurthy}, \bibinfo{person}{Benoit Steiner},
  \bibinfo{person}{Lu Fang}, \bibinfo{person}{Junjie Bai}, {and}
  \bibinfo{person}{Soumith Chintala}.} \bibinfo{year}{2019}\natexlab{}.
\newblock \showarticletitle{PyTorch: An Imperative Style, High-Performance Deep
  Learning Library}.
\newblock In \bibinfo{booktitle}{\emph{Advances in Neural Information
  Processing Systems 32}}, \bibfield{editor}{\bibinfo{person}{H.~Wallach},
  \bibinfo{person}{H.~Larochelle}, \bibinfo{person}{A.~Beygelzimer},
  \bibinfo{person}{F.~d\textquotesingle Alch\'{e}-Buc},
  \bibinfo{person}{E.~Fox}, {and} \bibinfo{person}{R.~Garnett}} (Eds.).
  \bibinfo{publisher}{Curran Associates, Inc.}, \bibinfo{pages}{8024--8035}.
\newblock
\urldef\tempurl%
\url{http://papers.neurips.cc/paper/9015-pytorch-an-imperative-style-high-performance-deep-learning-library.pdf}
\showURL{%
\tempurl}


\bibitem[\protect\citeauthoryear{Pedregosa, Varoquaux, Gramfort, Michel,
  Thirion, Grisel, Blondel, Prettenhofer, Weiss, Dubourg, Vanderplas, Passos,
  Cournapeau, Brucher, Perrot, and Duchesnay}{Pedregosa et~al\mbox{.}}{2011}]%
        {scikit-learn}
\bibfield{author}{\bibinfo{person}{F. Pedregosa}, \bibinfo{person}{G.
  Varoquaux}, \bibinfo{person}{A. Gramfort}, \bibinfo{person}{V. Michel},
  \bibinfo{person}{B. Thirion}, \bibinfo{person}{O. Grisel},
  \bibinfo{person}{M. Blondel}, \bibinfo{person}{P. Prettenhofer},
  \bibinfo{person}{R. Weiss}, \bibinfo{person}{V. Dubourg}, \bibinfo{person}{J.
  Vanderplas}, \bibinfo{person}{A. Passos}, \bibinfo{person}{D. Cournapeau},
  \bibinfo{person}{M. Brucher}, \bibinfo{person}{M. Perrot}, {and}
  \bibinfo{person}{E. Duchesnay}.} \bibinfo{year}{2011}\natexlab{}.
\newblock \showarticletitle{Scikit-learn: Machine Learning in {P}ython}.
\newblock \bibinfo{journal}{\emph{Journal of Machine Learning Research}}
  \bibinfo{volume}{12} (\bibinfo{year}{2011}), \bibinfo{pages}{2825--2830}.
\newblock


\bibitem[\protect\citeauthoryear{Scarselli, Gori, Tsoi, Hagenbuchner, and
  Monfardini}{Scarselli et~al\mbox{.}}{2008}]%
        {Scarselli::2008}
\bibfield{author}{\bibinfo{person}{Franco Scarselli}, \bibinfo{person}{Marco
  Gori}, \bibinfo{person}{Ah~Chung Tsoi}, \bibinfo{person}{Markus
  Hagenbuchner}, {and} \bibinfo{person}{Gabriele Monfardini}.}
  \bibinfo{year}{2008}\natexlab{}.
\newblock \showarticletitle{The graph neural network model}.
\newblock \bibinfo{journal}{\emph{IEEE transactions on neural networks}}
  \bibinfo{volume}{20}, \bibinfo{number}{1} (\bibinfo{year}{2008}),
  \bibinfo{pages}{61--80}.
\newblock


\bibitem[\protect\citeauthoryear{Sherstinsky}{Sherstinsky}{2018}]%
        {rnn2}
\bibfield{author}{\bibinfo{person}{Alex Sherstinsky}.}
  \bibinfo{year}{2018}\natexlab{}.
\newblock \showarticletitle{Fundamentals of Recurrent Neural Network {(RNN)}
  and Long Short-Term Memory {(LSTM)} Network}.
\newblock \bibinfo{journal}{\emph{CoRR}}  \bibinfo{volume}{abs/1808.03314}
  (\bibinfo{year}{2018}).
\newblock
\showeprint[arxiv]{1808.03314}
\urldef\tempurl%
\url{http://arxiv.org/abs/1808.03314}
\showURL{%
\tempurl}


\bibitem[\protect\citeauthoryear{Smith and Topin}{Smith and Topin}{2017}]%
        {onecycle}
\bibfield{author}{\bibinfo{person}{Leslie~N. Smith} {and}
  \bibinfo{person}{Nicholay Topin}.} \bibinfo{year}{2017}\natexlab{}.
\newblock \showarticletitle{Super-Convergence: Very Fast Training of Residual
  Networks Using Large Learning Rates}.
\newblock \bibinfo{journal}{\emph{CoRR}}  \bibinfo{volume}{abs/1708.07120}
  (\bibinfo{year}{2017}).
\newblock
\showeprint[arXiv]{1708.07120}
\urldef\tempurl%
\url{http://arxiv.org/abs/1708.07120}
\showURL{%
\tempurl}


\bibitem[\protect\citeauthoryear{Srivastava, Hinton, Krizhevsky, Sutskever, and
  Salakhutdinov}{Srivastava et~al\mbox{.}}{2014}]%
        {Srivastava2014}
\bibfield{author}{\bibinfo{person}{Nitish Srivastava},
  \bibinfo{person}{Geoffrey Hinton}, \bibinfo{person}{Alex Krizhevsky},
  \bibinfo{person}{Ilya Sutskever}, {and} \bibinfo{person}{Ruslan
  Salakhutdinov}.} \bibinfo{year}{2014}\natexlab{}.
\newblock \showarticletitle{Dropout: A Simple Way to Prevent Neural Networks
  from Overfitting}.
\newblock \bibinfo{journal}{\emph{J. Mach. Learn. Res.}} \bibinfo{volume}{15},
  \bibinfo{number}{1} (\bibinfo{date}{Jan.} \bibinfo{year}{2014}),
  \bibinfo{pages}{1929–1958}.
\newblock
\showISSN{1532-4435}


\bibitem[\protect\citeauthoryear{Sundararajan, Taly, and Yan}{Sundararajan
  et~al\mbox{.}}{2017}]%
        {integrated-grad}
\bibfield{author}{\bibinfo{person}{Mukund Sundararajan}, \bibinfo{person}{Ankur
  Taly}, {and} \bibinfo{person}{Qiqi Yan}.} \bibinfo{year}{2017}\natexlab{}.
\newblock \showarticletitle{Axiomatic Attribution for Deep Networks}.
\newblock \bibinfo{journal}{\emph{CoRR}}  \bibinfo{volume}{abs/1703.01365}
  (\bibinfo{year}{2017}).
\newblock
\showeprint[arXiv]{1703.01365}
\urldef\tempurl%
\url{http://arxiv.org/abs/1703.01365}
\showURL{%
\tempurl}


\bibitem[\protect\citeauthoryear{van~der Maaten and Hinton}{van~der Maaten and
  Hinton}{2008}]%
        {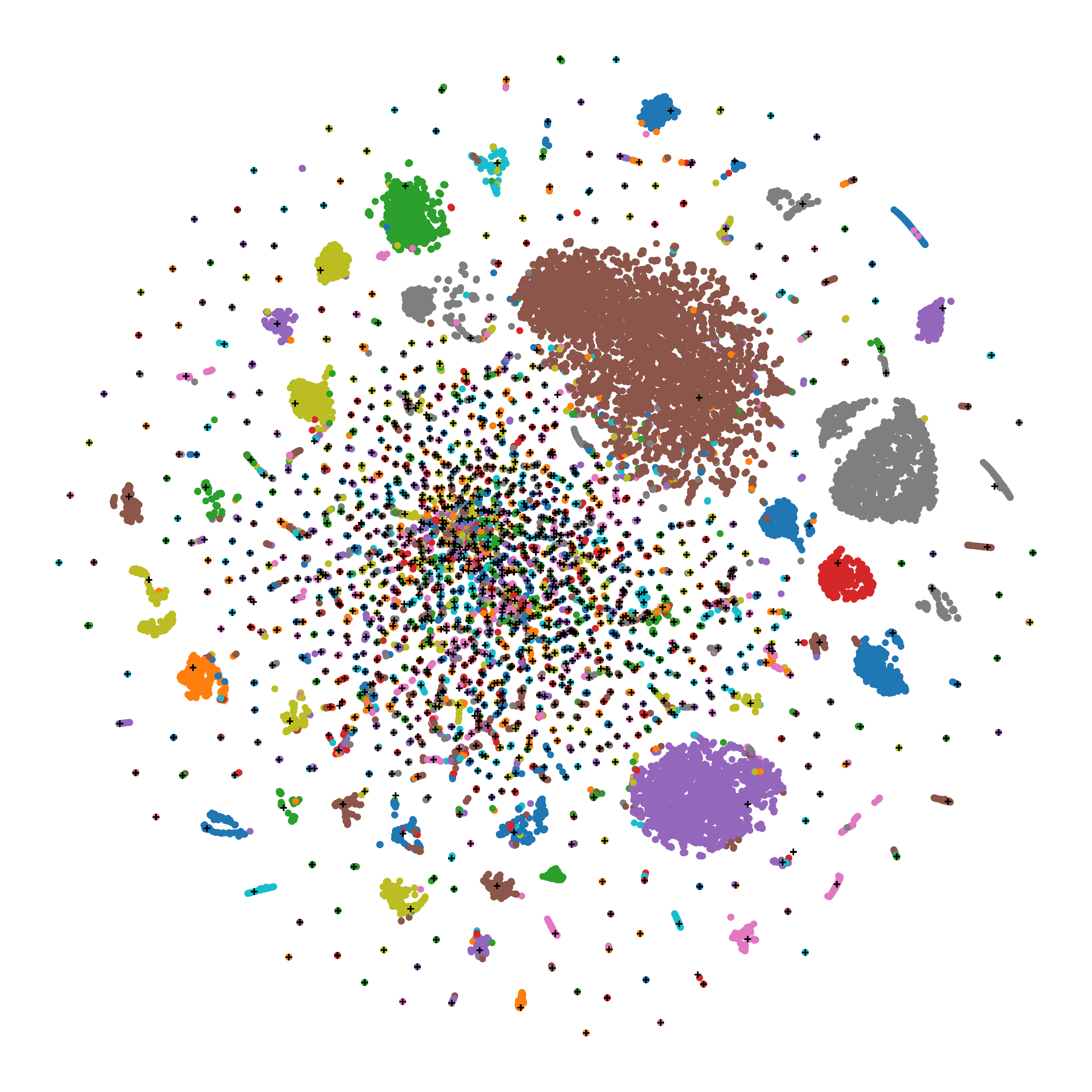}
\bibfield{author}{\bibinfo{person}{Laurens van~der Maaten} {and}
  \bibinfo{person}{Geoffrey Hinton}.} \bibinfo{year}{2008}\natexlab{}.
\newblock \showarticletitle{Visualizing Data using {t-SNE}}.
\newblock \bibinfo{journal}{\emph{Journal of Machine Learning Research}}
  \bibinfo{volume}{9} (\bibinfo{year}{2008}), \bibinfo{pages}{2579--2605}.
\newblock
\urldef\tempurl%
\url{http://www.jmlr.org/papers/v9/vandermaaten08a.html}
\showURL{%
\tempurl}


\bibitem[\protect\citeauthoryear{Vaswani, Shazeer, Parmar, Uszkoreit, Jones,
  Gomez, Kaiser, and Polosukhin}{Vaswani et~al\mbox{.}}{2017}]%
        {Transformers::2017}
\bibfield{author}{\bibinfo{person}{Ashish Vaswani}, \bibinfo{person}{Noam
  Shazeer}, \bibinfo{person}{Niki Parmar}, \bibinfo{person}{Jakob Uszkoreit},
  \bibinfo{person}{Llion Jones}, \bibinfo{person}{Aidan~N Gomez},
  \bibinfo{person}{{\L}ukasz Kaiser}, {and} \bibinfo{person}{Illia
  Polosukhin}.} \bibinfo{year}{2017}\natexlab{}.
\newblock \showarticletitle{Attention is all you need}. In
  \bibinfo{booktitle}{\emph{Advances in neural information processing
  systems}}. \bibinfo{pages}{5998--6008}.
\newblock


\bibitem[\protect\citeauthoryear{Wang, Elworth, Liu, and Treangen}{Wang
  et~al\mbox{.}}{2019a}]%
        {Wang::2019}
\bibfield{author}{\bibinfo{person}{Qi Wang}, \bibinfo{person}{RA Elworth},
  \bibinfo{person}{Tian~Rui Liu}, {and} \bibinfo{person}{Todd~J Treangen}.}
  \bibinfo{year}{2019}\natexlab{a}.
\newblock \showarticletitle{Faster pan-genome construction for efficient
  differentiation of naturally occurring and engineered plasmids with plaster}.
  In \bibinfo{booktitle}{\emph{19th International Workshop on Algorithms in
  Bioinformatics (WABI 2019)}}. Schloss Dagstuhl-Leibniz-Zentrum fuer
  Informatik.
\newblock


\bibitem[\protect\citeauthoryear{Wang, Kille, Liu, Elworth, and Treangen}{Wang
  et~al\mbox{.}}{2021}]%
        {Wang2021}
\bibfield{author}{\bibinfo{person}{Qi Wang}, \bibinfo{person}{Bryce Kille},
  \bibinfo{person}{Tian~Rui Liu}, \bibinfo{person}{R.~A.~Leo Elworth}, {and}
  \bibinfo{person}{Todd~J. Treangen}.} \bibinfo{year}{2021}\natexlab{}.
\newblock \showarticletitle{PlasmidHawk improves lab of origin prediction of
  engineered plasmids using sequence alignment}.
\newblock \bibinfo{journal}{\emph{Nature Communications}} \bibinfo{volume}{12},
  \bibinfo{number}{1} (\bibinfo{date}{26 Feb} \bibinfo{year}{2021}),
  \bibinfo{pages}{1167}.
\newblock
\showISSN{2041-1723}
\urldef\tempurl%
\url{https://doi.org/10.1038/s41467-021-21180-w}
\showDOI{\tempurl}


\bibitem[\protect\citeauthoryear{Wang, Zhang, Huang, and Scott}{Wang
  et~al\mbox{.}}{2019b}]%
        {Wang2019}
\bibfield{author}{\bibinfo{person}{Xun Wang}, \bibinfo{person}{Haozhi Zhang},
  \bibinfo{person}{Weilin Huang}, {and} \bibinfo{person}{Matthew~R. Scott}.}
  \bibinfo{year}{2019}\natexlab{b}.
\newblock \showarticletitle{Cross-Batch Memory for Embedding Learning}.
\newblock \bibinfo{journal}{\emph{CoRR}}  \bibinfo{volume}{abs/1912.06798}
  (\bibinfo{year}{2019}).
\newblock
\showeprint[arxiv]{1912.06798}
\urldef\tempurl%
\url{http://arxiv.org/abs/1912.06798}
\showURL{%
\tempurl}


\bibitem[\protect\citeauthoryear{Wang, Yao, Kwok, and Ni}{Wang
  et~al\mbox{.}}{2020}]%
        {Wang2020}
\bibfield{author}{\bibinfo{person}{Yaqing Wang}, \bibinfo{person}{Quanming
  Yao}, \bibinfo{person}{James~T. Kwok}, {and} \bibinfo{person}{Lionel~M. Ni}.}
  \bibinfo{year}{2020}\natexlab{}.
\newblock \showarticletitle{Generalizing from a Few Examples}.
\newblock \bibinfo{journal}{\emph{Comput. Surveys}} \bibinfo{volume}{53},
  \bibinfo{number}{3} (\bibinfo{date}{July} \bibinfo{year}{2020}),
  \bibinfo{pages}{1--34}.
\newblock
\urldef\tempurl%
\url{https://doi.org/10.1145/3386252}
\showDOI{\tempurl}


\bibitem[\protect\citeauthoryear{Xu and Tian}{Xu and Tian}{2015}]%
        {clustering2}
\bibfield{author}{\bibinfo{person}{Dongkuan Xu} {and} \bibinfo{person}{Yingjie
  Tian}.} \bibinfo{year}{2015}\natexlab{}.
\newblock \showarticletitle{A Comprehensive Survey of Clustering Algorithms}.
\newblock  \bibinfo{volume}{2}, \bibinfo{number}{2} (\bibinfo{date}{June}
  \bibinfo{year}{2015}), \bibinfo{pages}{165--193}.
\newblock
\urldef\tempurl%
\url{https://doi.org/10.1007/s40745-015-0040-1}
\showDOI{\tempurl}


\bibitem[\protect\citeauthoryear{Yu, Gao, Rong, Song, and Xiong}{Yu
  et~al\mbox{.}}{2017}]%
        {recsys2}
\bibfield{author}{\bibinfo{person}{Junliang Yu}, \bibinfo{person}{Min Gao},
  \bibinfo{person}{Wenge Rong}, \bibinfo{person}{Yuqi Song}, {and}
  \bibinfo{person}{Qingyu Xiong}.} \bibinfo{year}{2017}\natexlab{}.
\newblock \showarticletitle{A Social Recommender Based on Factorization and
  Distance Metric Learning}.
\newblock   \bibinfo{volume}{5} (\bibinfo{year}{2017}),
  \bibinfo{pages}{21557--21566}.
\newblock
\urldef\tempurl%
\url{https://doi.org/10.1109/access.2017.2762459}
\showDOI{\tempurl}


\bibitem[\protect\citeauthoryear{Zou, Xing, Wei, and Liu}{Zou
  et~al\mbox{.}}{2018}]%
        {Zou2018}
\bibfield{author}{\bibinfo{person}{Quan Zou}, \bibinfo{person}{Pengwei Xing},
  \bibinfo{person}{Leyi Wei}, {and} \bibinfo{person}{Bin Liu}.}
  \bibinfo{year}{2018}\natexlab{}.
\newblock \showarticletitle{Gene2vec: gene subsequence embedding for prediction
  of mammalian N6-methyladenosine sites from {mRNA}}.
\newblock \bibinfo{journal}{\emph{{RNA}}} \bibinfo{volume}{25},
  \bibinfo{number}{2} (\bibinfo{date}{Nov.} \bibinfo{year}{2018}),
  \bibinfo{pages}{205--218}.
\newblock
\urldef\tempurl%
\url{https://doi.org/10.1261/rna.069112.118}
\showDOI{\tempurl}


\end{thebibliography}


\end{document}